%% file: arxiv.tex
\documentclass{article} % For LaTeX2e
\usepackage{arxiv,times}

\input{math_commands.tex}

\usepackage{hyperref}
\usepackage{url}
\usepackage{enumitem}
\usepackage{graphicx}
\usepackage{cleveref}
\usepackage{algorithm}
\usepackage{algorithmic}
\usepackage{subcaption}
\usepackage{booktabs}
\usepackage{multirow}

\usepackage{amsthm}
\theoremstyle{plain}
\newtheorem{theorem}{Theorem}[section]

\def\algo{{\textsc{LoRAct}}}

\title{Memory-Efficient Fine-Tuning via Low-Rank Activation Compression}

\author{%
Jiang-Xin~Shi$^1$ \;
Wen-Da Wei$^1$ \;
Jin-Fei Qi$^1$ \;
Xuanyu Chen$^1$ \;
Tong~Wei$^2$ \;
Yu-Feng~Li$^1$\thanks{Corresponding author.} \\
$^1$Nanjing University \;
$^2$Southeast University\\
\texttt{\{shijx,liyf\}@lamda.nju.edu.cn}
}

\iclrfinalcopy % Uncomment for camera-ready version, but NOT for submission.
\begin{document}

\maketitle

\begin{abstract}
The parameter-efficient fine-tuning paradigm has garnered significant attention with the advancement of foundation models. Although numerous methods have been proposed to reduce the number of trainable parameters, their substantial memory overhead remains a critical bottleneck that hinders practical deployment. In this paper, we observe that model activations constitute a major source of memory consumption, especially under large batch sizes and long context lengths; however, the rank of the activations remains consistently low. Motivated by this insight, we propose a memory-efficient fine-tuning approach \textit{Low-Rank Activation Compression} (\algo). Unlike prior work, \algo\ provides a more flexible and versatile compressing strategy that can be applied online during the forward pass without the need for any calibration data. Moreover, \algo\ incorporates a novel sampling-based orthogonal decomposition algorithm specifically designed for low-rank matrices, offering improved computational efficiency and a tighter error bound compared to the widely used RSVD. Experiments on both vision and language tasks demonstrate the effectiveness of \algo. Notably, \algo\ further reduces activation memory by approximately 80\% in comparison with the widely adopted LoRA method, while maintaining competitive performance. The source code is available at \url{https://github.com/shijxcs/meft}.
\end{abstract}

\section{Introduction}

With the rapid development of foundation models, the parameter-efficient fine-tuning (PEFT) has emerged as a prevailing paradigm for adapting the models to downstream tasks~\citep{houlsby2019parameter, lester2021power, li2021prefix, hu2022lora, yang2024s}. The primary motivation of PEFT is to fine-tune foundation models using only a small set of trainable parameters. By this means, it aims to leverage minimal additional memory overhead while enhancing the adaptability to specific downstream tasks. However, despite reducing the number of trainable parameters, PEFT still incurs non-negligible memory consumption during fine-tuning, which significantly constrains its flexibility in practical scenarios~\citep{simoulin2025memory, huang2025sola}.

\begin{figure}[!t]
    \centering
    \begin{subfigure}{0.43\linewidth}
        \includegraphics[width=\linewidth]{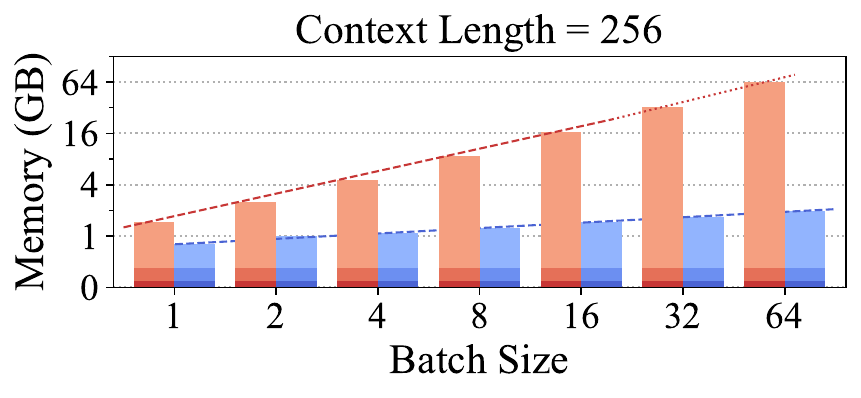}
    \end{subfigure}
    \begin{subfigure}{0.325\linewidth}
        \includegraphics[width=\linewidth]{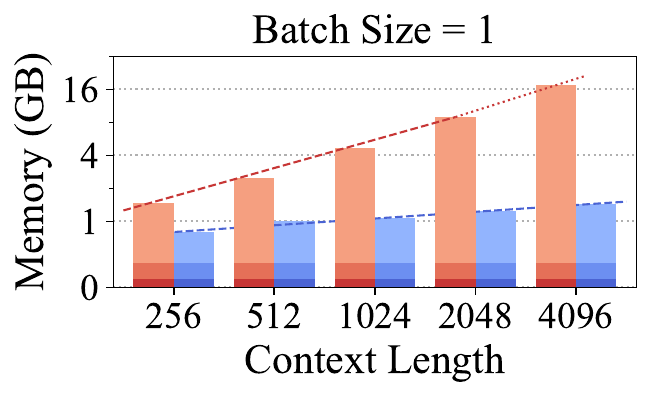}
    \end{subfigure}
    \begin{subfigure}{0.21\linewidth}
        \includegraphics[width=\linewidth,trim=-10 0 10 0]{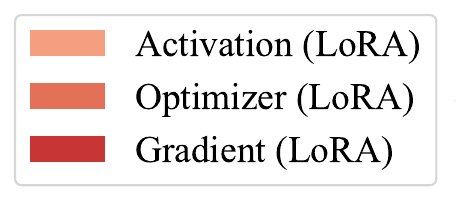} \\
        \includegraphics[width=\linewidth,trim=-10 -15 10 0]{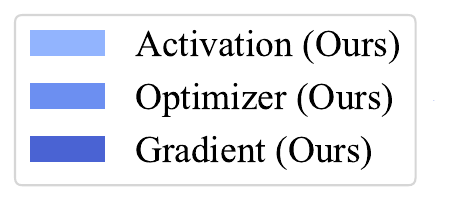}
    \end{subfigure}
    \vspace{-0.1in}
  \setlength{\belowcaptionskip}{-0.1in}
    \caption{The additional memory cost for fine-tuning LLaMa2-7B with LoRA. The endpoints of the dashed line denote estimated results, as these configurations are infeasible on consumer-grade GPU.}
    \label{fig:memory_cost}
\end{figure}

In this paper, we systematically investigate the underlying reasons of the substantial memory consumption of the PEFT method. As illustrated in \Cref{fig:memory_cost}, the memory costs associated with gradients and optimizer states are significantly reduced, which is owing to the reduction in the number of trainable parameters. However, the model activation memory constitutes the dominant factor of the total memory overhead, which scales approximately linearly with both batch size and context length. The substantial memory consumption is inconsistent with the original motivation of PEFT, which inevitably limits the practical deployment. Therefore, it is crucial to develop methods to advance parameter-efficient fine-tuning towards a memory-efficient process.

Prevailing memory-efficient fine-tuning methods primarily focus on the compression of gradients and optimizer states, such as Galore~\citep{zhao2024galore}, FLora~\citep{DBLP:conf/icml/HaoCM24}, and Apollo~\citep{zhu2024apollo}. However, these approaches are ill-suited for the PEFT setting, since PEFT already drastically reduces the number of trainable parameters, and the memory overhead from gradients and optimizer states is mostly manageable. Moreover, these methods neglect the problem of substantial activation memory. Despite some prior work has explored activation compression, these methods still suffer from multiple limitations. For instance, LoRA-FA~\citep{zhang2023lora} and CompAct~\citep{shamshoum2024compact} provide compression by confining the scope to specific layers, which may fail to address the full memory overhead. CoLA~\citep{liu2025cola} and VeLoRA~\citep{miles2024velora} employ sub-optimal compression algorithms that may significantly degrade performance. ESPACE~\citep{sakr2024espace} and SoLA~\citep{huang2025sola} require additional calibration data for compression, which introduces extra storage overhead and diminishes their practical utility.

To address this critical bottleneck, we propose a novel memory-efficient fine-tuning approach named \textbf{Lo}w-\textbf{R}ank \textbf{Act}ivation Compression (\algo), which aims to reduce activation memory overhead while maintaining fine-tuning performance. By leveraging the inherent low-rank characteristic of model activation, \algo\ employs an online compression mechanism during the forward pass to reduce the activation memory consumption, eliminating the requirement for any additional calibration data. To mitigate the computational overhead of the online compression, we introduce a novel sampling-based orthogonal decomposition algorithm that substantially enhances computational efficiency and decomposition quality. Furthermore, to facilitate the adaptability to the widely adopted Transformer-based foundation models, we design a tailored pre-norm activation compression strategy that ensures both compatibility and improved performance. To verify the effectiveness of \algo, we conduct experiments on both language and vision tasks. The results show that \algo\ consistently reduces activation memory while maintaining fine-tuning performance. Notably, by controlling the compression ratio,  \algo\ achieves competitive performance compared to LoRA while saving approximately 80\% of the memory across all experiments.

The main contributions of this paper are summarized as follows:
\begin{itemize}[topsep=0in,leftmargin=0.2in,itemindent=0in,itemsep=0in]
    \item We empirically reveal the inherent low-rank nature of activations in the context of PEFT, thus validating the feasibility of activation compression.
    \item We propose \algo, a flexible and versatile online activation compression strategy that can be applied to any differentiable functions and is free of any calibration data.
    \item To ensure computational efficiency, we introduce a novel sampling-based orthogonal decomposition algorithm that reduces computational overhead while maintaining theoretical guarantees.
    \item We design a tailored pre-norm compression mechanism that can be seamlessly integrated into Transformer-based models. The proposed method achieves approximately 80\% of activation memory reduction compared to LoRA, while maintaining competitive performance.
\end{itemize}

\section{Background}
\subsection{Is Parameter-Efficient equivalent to Memory-Efficient?}

The parameter-efficient fine-tuning paradigm optimizes only a small subset of parameters while exhibiting remarkable performance, which has attracted considerable research interest~\citep{houlsby2019parameter,hu2022lora,ding2023parameter,han2024parameter}. The learning objective of parameter-efficient fine-tuning can be formulated as
\begin{equation}
    L=\Ls\left(f(\mX;\vtheta,\vtheta_0),\mY\right)
\end{equation}
where $\vtheta$ denotes the trainable parameters, $\vtheta_0$ represents the fixed parameters, and $||\vtheta||_0\ll||\vtheta_0||_0$. During the backward pass, the trainable parameters $\vtheta$ are updated along the negative gradient direction, i.e., $\nabla_\vtheta L$, using the backpropagation algorithm~\citep{rumelhart1986learning}. However, the overall memory requirement is not solely determined by the number of trainable parameters. Specifically, memory consumption during fine-tuning arises mainly from the following four components:

\begin{itemize}[topsep=0in,leftmargin=0.17in,itemindent=0in,itemsep=0.02in]
\item \textit{Model parameters}, which comprises both trainable parameters $\vtheta$ and fixed parameters $\vtheta_0$. The memory usage is related to the quantity of model parameters and storage format.
\item \textit{Gradients} of trainable parameters $\vg=\nabla_\vtheta L$, which are stored for parameter optimization. The memory consumption of gradients is equal to that of trainable parameters.
\item \textit{Optimizer states} for improving optimization, such as SGD momentum~\citep{sutskever2013importance} and Adam~\citep{kingma2014adam}. The memory usage is also related to the trainable parameters.
\item \textit{Activations} for gradient computation, which are computed during the forward pass and retained for use in the backward pass. For example, in attention mechanism~\citep{bahdanau2015neural}, the inputs $\mQ$, $\mK$, $\mV$, and the attention score $\mathrm{Softmax}(\mQ\mK^\top/\sqrt{d})$ are stored to facilitate backpropagation. Similarly, for nonlinear functions such as $\va=\mathrm{Softmax}(\vx)$ and $\va=\mathrm{Sigmoid}(\vx)$, the output $\va$ is preserved for computing the partial derivative $\partial\va/\partial\vx$. In practice, the memory of activations is dependent on the model architecture.
\end{itemize}

In parameter-efficient fine-tuning, the number of trainable parameters is significantly smaller than the total number of model parameters. As a result, the memory required to store gradients and optimizer states is substantially reduced. However, the consumption of activation memory remains non-negligible. Although freezing model weights eliminates the need to store activations for the corresponding linear layers, a considerable amount of activation memory is still required for nonlinear layers during backpropagation, such as the normalization layers, multi-head self-attention layers, and multi-layer perceptrons. Consequently, activation memory emerges as a dominant factor in overall memory consumption.

To evaluate the additional memory cost of fine-tuning, we perform a series of experiments on LLaMa2-7B~\citep{touvron2023llama} using LoRA~\citep{hu2022lora}. The results are presented in \Cref{fig:memory_cost}. Although the number of trainable parameters is substantially reduced, leading to a total memory cost of less than 1 GB for gradients and optimizer states, the activation memory remains considerably high. Moreover, the activation memory increases linearly with both batch size and context length. In some configurations, the memory demand can reach up to 64 GB, which even exceeds the total memory capacity of consumer-grade GPUs. Such considerable memory requirements make large batch sizes or context lengths infeasible in practical deployments.

\subsection{Is Activation Memory Compressible?}

\begin{figure}[!t]
    \centering
    \begin{subfigure}{0.325\linewidth}
        \includegraphics[width=\linewidth]{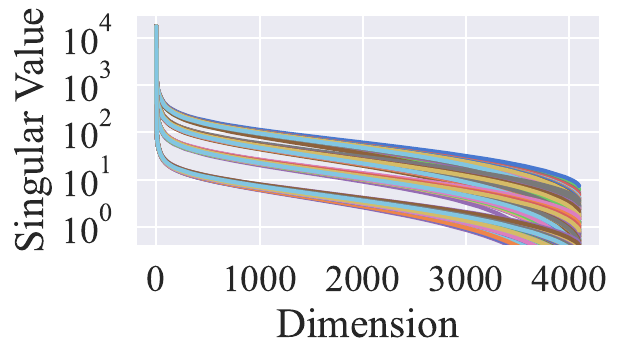}
        \setlength{\abovecaptionskip}{-0.15in}
        \caption{}
        \label{fig:singular_value}
    \end{subfigure}
    \begin{subfigure}{0.325\linewidth}
        \includegraphics[width=\linewidth]{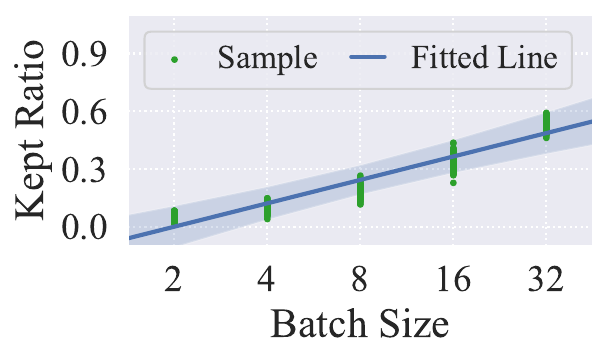}
        \setlength{\abovecaptionskip}{-0.15in}
        \caption{}
        \label{fig:kept_ratio_batch_size}
    \end{subfigure}
    \begin{subfigure}{0.325\linewidth}
        \includegraphics[width=\linewidth]{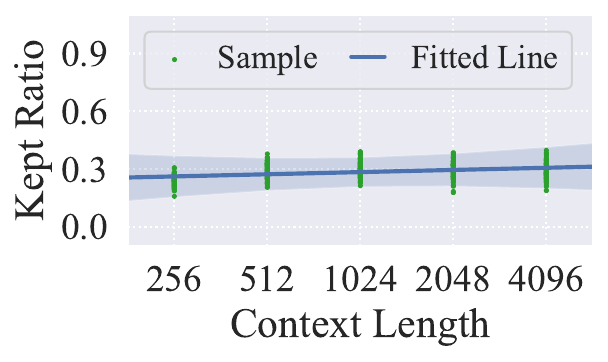}
        \setlength{\abovecaptionskip}{-0.15in}
        \caption{}
        \label{fig:kept_ratio_context_length}
    \end{subfigure}
    \vspace{-0.1in}
  \setlength{\belowcaptionskip}{-0.1in}
    \caption{(a) The singular value distributions of model activations from different layers and batches. (b-c) The ratio of kept dimensions when 90\% of energy is retained.}
    \label{fig:activation_rank}
\end{figure}

The high activation memory overhead in parameter-efficient fine-tuning poses a significant challenge to its objective of achieving lightweight memory consumption. This naturally raises a question: \textit{Can activation memory be effectively compressed?}

To explore this question, we first extract the activations from samples in the Alpaca dataset~\citep{taori2023stanford} using the LLaMa2-7B model~\citep{touvron2023llama}. We then analyze the singular values of these activation across different sample batches and model layers. The results illustrated in \Cref{fig:singular_value} indicate that the singular values consistently exhibit a long-tailed distribution regardless of the source of activation. Notably, the largest singular value exceeds $10^4$, while most of the remaining values fall within the range of $10^0$ to $10^2$. This suggests that the activation matrices are essentially low-rank, with only a few dimensions capturing the majority of the information, whereas the rest contribute only marginally.

To further investigate the potential of activation compression under large batch sizes or context lengths, we further analyze the characteristics of activation with increased sizes. To quantify the low-rank nature of activations, we compute the ratio of kept dimensions when 90\% of energy (i.e., the sum of singular values) is retained. The results are reported in \Cref{fig:kept_ratio_batch_size,fig:kept_ratio_context_length}. For small batch sizes of 2 or 4, only about 10\% of the dimensions need to be retained to preserve 90\% of the total energy. Even at a large batch size of 32, the kept ratio remains around 50\%, which indicates significant potential for memory reduction via activation compression. Besides, increasing the context length has only a minor effect on the rank of activations, with the kept ratio stabilizing at approximately 30\%, which further supports the feasibility of compressing activation sizes.

Some prior research has also explored the potential of activation compression, which however, is either unpractical or less accurate. For example, 
LoRA-FA~\citep{zhang2023lora} freezes the down projection and optimizes only the up projection in each LoRA layer; however, it can only reduce limited size of activations. 
CompAct~\citep{shamshoum2024compact} projects the activation of each linear layer into a low-rank subspace, which is incompatible with the prevailing parameter-efficient fine-tuning, where the activations from linear layers are already eliminated. 
CoLA~\citep{liu2025cola} decomposes the full-size weight matrix into low-rank matrices, which inevitably harms the original inference capability.
ESPACE~\citep{sakr2024espace} projects the activation using a static matrix, which loses flexibility when facing diverse activations.
SoLA~\citep{huang2025sola} conducts low-rank decomposition approach based on an additional calibration dataset, which is unfeasible in most practical scenarios.
To overcome the limitations of existing works, we propose \algo, a flexible and versatile compressing method that facilitates memory-efficient fine-tuning.

\section{Proposed Method}

\subsection{\algo: Low-Rank Activation Compression}
\label{sec:framework}

\begin{figure}[!t]
  \setlength{\abovecaptionskip}{0.1in}
    \centering
    \includegraphics[width=0.9\linewidth]{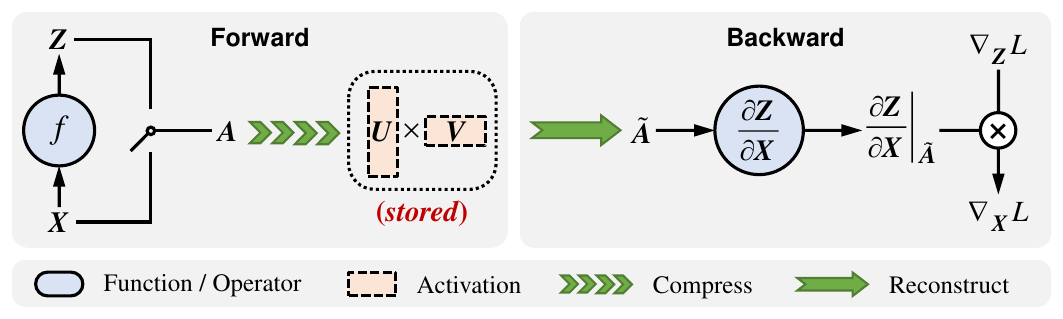}
  \setlength{\belowcaptionskip}{-0.1in}
    \caption{An illustration of \algo.}
    \label{fig:loract}
\end{figure}

We propose a memory-efficient fine-tuning method termed \textit{Low-Rank Activation Compression} (\algo). The overall framework is illustrated in \Cref{fig:loract}. Specifically, \algo\ first calculates the function output during the forward pass, and then extracts and decomposes the activation into low-rank approximations:
\begin{equation}
    \mZ=f\left(\mX\right)
\end{equation}
\begin{equation}
    \mU,\mV=\mathrm{Decompose}\left(\mA,k\right)
\end{equation}
where $\mA\in\sR^{m\times n}$ is the activation for computing $\partial\mZ/\partial\mX$, which typically comes from $\mX$ or $\mZ$. \algo\ applies low-rank matrix decomposition on $\mA$ to obtain $\mA\approx\mU\mV$, where $\mU\in\sR^{m\times k}$ and $\mV\in\sR^{k\times n}$ are two low-rank matrices. During the backward pass, \algo\ first reconstructs the activation by $\tilde\mA=\mU\mV$, and then utilize $\tilde\mA$ to calculate the partial derivative as well as the gradient:
\begin{equation}
    \tilde\mA=\mathrm{Reconstruct}\left(\mU,\mV\right)
\end{equation}
\begin{equation}
    \nabla_\mX L=\nabla_\mZ L\cdot\frac{\partial\mZ}{\partial\mX}\bigg{|}_{\tilde{\mA}}
\end{equation}
By compressing and storing only the low-rank matrices, \algo\ reduces the activation memory from $\gO(mn)$ to $\gO((m+n)k)$, where $k\ll m,n$. 
Furthermore, by incorporating an online compression mechanism, \algo\ adaptively compresses activations into their corresponding optimal components. This process is agnostic to the diverse properties of activations and eliminates the need for calibration data.

It is worth noting that \algo\ can be applied to any differentiable functions and modules. In contrast, existing activation compression methods only consider a single type of function, such as compressing linear functions~\citep{shamshoum2024compact} or normalization layers~\citep{hu2025quad}, which substantially limits their generalizability. Moreover, prior methods~\citep{yu2023compressing,sakr2024espace,liu2025cola} compress activations at the expense of simplifying base modules or intermediate variables, which inevitably constrains the inherent capability of the base model. In contrast, \algo\ does not compress any modules during the forward process, thus ensuring the basic inference performance of the model.

We prove that compressing activations during the forward pass, which is a common practice in many existing approaches, will lead to an accumulation of error with respect to the learning objective.
\begin{theorem}  \label{4.0}
Suppose there are $N$ compressed activation matrices with the $i$-th compression incurring an error $\sigma_i$.
Between consecutive activations there are fusion operators $\{f_i\}_{i=1}^{N-1}$, each being Lipschitz continuous with constant $L_i$.
Let $L=\max_{i\in[N-1]}{L_i}$.
For any input $\vx$ and label $\vy$, we have:
$$\mathcal{L}(F(\vx),\vy)-\mathcal{L}(F_{comp}(\vx),\vy) \leq \sqrt{2}L_{\mW}\Sigma_{i=1}^{N}(L^{N-i}\sigma_i)$$
where $F$ denotes the overall forward mapping of the model, $L_{\mW}$is the Lipschitz constant of the final linear layer,
$\mathcal{L}$ denotes the cross-entropy loss function.
\end{theorem}

\subsection{The Enhanced Decomposition Algorithm}
\label{sec:decomposition}

\algo\ employs an online matrix compression mechanism during the forward pass, and the effectiveness and efficiency of the algorithm are largely determined by this mechanism.
In general, a matrix decomposition method approximates a large matrix $\mA_{m \times n}$ by the product of two rank-$k$ matrices $\mU_{m\times k}$ and $\mV_{k \times n}$, thereby reducing the storage overhead. 

Some previous work~\citep{DBLP:conf/icml/HaoCM24} attempts to approximate $\mA$ by projecting its image into a lower-dimensional subspace via a random projection. 
Specifically, given a Gaussian matrix $\mG_{l \times m}$, one defines $\mU=\tfrac{1}{l}\mG^\top$ and $\mV=\mG\mA$, and approximates $\mA$ by $\tfrac{1}{l}\mG^\top \mG \mA$. However, this approximation can incur a large error and we provide a lower bound on its  error.
\begin{theorem} \label{4.1}
Let $\mG_{l\times m}$ be a standard Gaussian matrix. The expected deviation admits the approximate lower bound: 
$$\mathbb{E}||\frac{1}{l}\mG^\top\mG\mA-\mA||\geq\sqrt{\frac{m-l}{m+1}}||\mA||$$
\end{theorem}
For better memory compression, it is typical to set $l \ll m$, in which case the lower bound is approximately $||\mA||$.
On the other hand, the error upper bound of Randomized SVD (RSVD) is $(1+\sqrt{\tfrac{k}{l-k}})\sigma_{k+1}(\mA) + \tfrac{e\sqrt{l}}{l-k}\sqrt{\sum_{j>k}\sigma_j^2}$~\citep{DBLP:journals/siamrev/HalkoMT11}.
Consequently, we establishe that RSVD outperforms random projection by a substantial margin, suggesting that approximate SVD methods constitute a more advantageous strategy.

\textit{Remark} 3.1. \
\textit{The conclusion we obtain is highly intuitive. This is because the RSVD method employs a Gaussian matrix to randomly approximate the column space of $\mA$, whereas random projection instead approximates the entire space $\mathbb{R}^m$. Clearly, approximating the column space of $\mA$ is an easier task, since $\mA$ may be rank-deficient or even low-rank.}

While Truncated SVD provides the optimal rank-$k$ approximation~\citep{eckart1936approximation}, its computational cost is prohibitive at $\mathcal{O}(mn \min(m,n))$.
RSVD reduce the complexity to $\mathcal{O}(mnk)$ and offer tight theoretical guarantees on approximation error, making them widely adopted in practice~\citep{DBLP:journals/siamrev/HalkoMT11}.
However, in large-scale models, applying RSVD to compress activations still incurs substantial computational overhead, making it difficult to meet the efficiency requirements of \algo.
To overcome this limitation, we introduce a novel decomposition algorithm.

The computational cost of RSVD mainly arises from three components: generating the standard Gaussian matrix, performing the QR decomposition, and executing the matrix multiplications. Notably, the Nyström method~\citep{JMLR:v6:drineas05a} does not require generating a Gaussian matrix, which motivates us to propose a Sampling-Based Orthogonal Decomposition algorithm that replaces the Gaussian matrix with a uniformly sampled submatrix $\mA_k^\top$, while keeping the rest of the algorithm identical to RSVD. This design eliminates the need to generate a Gaussian matrix, thereby further reducing the computational overhead and combining the strengths of Nyström and RSVD. The detailed procedure of the algorithm is shown in Algorithm~\ref{alg:decompose_algo}.

\begin{algorithm}[t]
\renewcommand{\algorithmicrequire}{\textbf{Input:}}
\renewcommand{\algorithmicensure}{\textbf{Output:}}
\caption{\textsc{Sampling-based Orthogonal Decomposition}}
\label{alg:decompose_algo}
% \small
\begin{algorithmic}[1]
\REQUIRE 
Original matrix $\mA\in\sR^{m\times n}$, expected rank $k$, number of iterations $t$.
\ENSURE
Decomposed matrices $\mU\in\sR^{m\times k}$ and $\mV\in\sR^{k\times n}$.
\STATE Randomly sample $k$ rows of $\mA$, denoted as $\mA_k\in\sR^{k\times n}$.
\STATE Compute $\mY=\mA\mA_k^\top$.
\FOR{$i=1,\cdots,t$}
\STATE Compute QR decomposition of $\mY=\mQ\mR$, where $\mQ\in\sR^{m\times k}$.
\STATE Compute $\mY=\mA(\mA^\top\mQ)$.
\ENDFOR
\STATE Compute QR decomposition of $\mY=\mQ\mR$, where $\mQ\in\sR^{m\times k}$.
\STATE Form $\mU=\mQ$ and compute $\mV=\mQ^\top\mA$.
\end{algorithmic}
\end{algorithm}

To establish the effectiveness of our algorithm, we provide a theoretical upper bound on its approximation error. 
Similar to RSVD, our method employs a test matrix to perform random sampling on the column space of $\mA$, thereby approximately preserving it in a lower-dimensional subspace. 
Therefore, the classical result established by \citet{DBLP:journals/siamrev/HalkoMT11} in Theorem~\ref{4.2} is highly relevant, for it underpins all fundamental results on SVD approximation error using test matrices.

\begin{theorem} \label{4.2}
Suppose that  $\mA = \mU \mSigma \mV^\top=\mU  \begin{bmatrix} \mSigma_1 & \bm0 \\ \bm0 & \mSigma_2 \\ \bm0 & \bm0  \end{bmatrix} 
\begin{bmatrix} \mV_1^\top \\ \mV_2^\top \end{bmatrix} $ and $\mathbf{\Omega}$ is an arbitrary test matrix, where
$\mV_1$ is the submatrix consisting of the first $k$ columns of $\mV$ and $\mV_2$  is the submatrix consisting of the remaining $n-k$ columns of $\mV$. $\mSigma_2$ is the the diagonal matrix containing the last $n-k$ singular values of $\mA$.
We define that $\mY=\mA \bm{\Omega}$, $\mathbf{\Omega}_1=\mV_1^\top\bm{\Omega}$ and $\mathbf{\Omega}_2=\mV_2^\top\bm{\Omega}$.

If $\bm{\Omega}_1$ has full row rank, we have:
$|| \mA -\mP_{\mY}\mA ||^2\leq ||\mSigma_2||^2+||\mSigma_2\bm{\Omega}_2\bm{\Omega}_1^+||^2$,
where $\bm{\Omega}_1^+$ denotes the Moore-Penrose pseudoinverse of $\bm{\Omega}_1$,
$\mP_{\mY}$ denotes the orthogonal projector onto $\mathrm{range}(\mY)$. 
\end{theorem}

Theorem~\ref{4.2} gives a deterministic bound for the SVD approximation error using test matrix.
The proof of this theorem in \citet{DBLP:journals/siamrev/HalkoMT11} contains certain inaccuracies; hence, we restate the theorem here and present a corrected proof in the appendix. 
Based on this, we further derive the theoretical upper bound of the approximation error for our proposed method.

\begin{theorem}  \label{4.3}
   For a matrix $\mA_{m \times n}$, suppose that we perform uniform random row sampling to obtain a submatrix $\mA_l \in \R^{l \times n}$, 
   and let the test matrix $\mathbf{\Omega} := \mA_l^\top$.
   Then we have the expectation bound 
   $$\mathbb{E}||\mA-\mQ\mQ^T\mA|| \leq (1+C\sqrt{\frac{\mu_k k}{l}})\sigma_{k+1}(\mA)+k e^{\frac{1}{c}}||\mA||,$$
   where $C$ is constant and $\mu_k:=\frac{m}{k}\,\max_{1 \le i \le m} || (\mU_k)_{i,:} ||_2$. $\mU_k$ denotes the first $k$ columns of $\mU$.
\end{theorem}

In summary, Theorem~\ref{4.3} shows that the error upper bound of our sampling-based method is relatively tight, explicitly depending on the $k$-coherence $\mu_k$ of $\mA$ through the subspace structure of $\mU_k$.
This result indicates that our approach preserves the comparable theoretical approximation guarantees as RSVD. At the same time, by avoiding the generation of Gaussian test matrices, our method achieves reduced computational overhead without incurring additional approximation error,
which meets our requirement.
All the theorems are proved in detail in the Appendix.

\subsection{Adaptation to Transformer-based Models}
\label{sec:transformer}

\begin{figure}[!t]
  \setlength{\abovecaptionskip}{0.05in}
    \centering
    \includegraphics[width=0.9\linewidth]{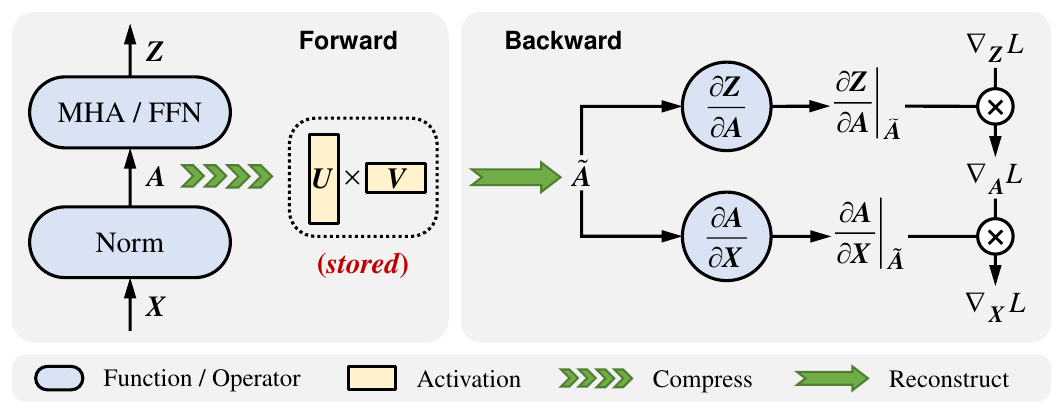}
  \setlength{\belowcaptionskip}{-0.1in}
    \caption{The adaptation of \algo\ for Transformer-based models. \algo\ leverages the convenience of the pre-norm structure to further save memory and computational costs.}
    \label{fig:loract-transformer}
\end{figure}

Transformer-based models~\citep{vaswani2017attention,dosovitskiy2021an} have been widely adopted in various scenarios, and serve as a fundamental architecture for many large-scale foundation models~\citep{touvron2023llama,liu2024deepseek}. Therefore, it is necessary to adapt \algo\ to the prevailing Transformer-based models.
One potential solution is to treat each Transformer layer as a black-box and directly compress the activation of input of each layer. However, such approach neglects the intrinsic architecture inside the layer, and may lead to redundant activation memory and unexpected estimation error.

To overcome this challenge, we introduce a \textit{pre-norm activation compression} strategy, which is tailored for the fundamental pre-norm unit in each Transformer layer. Formally, a Transformer layer can be expressed as
\begin{equation}
    \mZ=\gF\left(\mathrm{Norm}\left(\mX\right)\right)+\mX 
\end{equation}
\begin{equation}
    \mathrm{Norm}\left(\mX\right)=\frac{\mX}{\mathrm{RMS}(\mX)}\cdot\bm{\gamma}
\end{equation}
where $\gF$ represents a sub-layer, which can be either a Multi-Head Attention (MHA) or a Feed-Forward Network (FFN). $\mathrm{RMS}$ denotes root mean square~\citep{zhang2019root}, and $\bm{\gamma}$ is a scaling vector. This structure is referred to as a pre-norm unit~\citep{wang2019learning}.
In conventional fine-tuning frameworks, both $\mathrm{Norm}$ and $\gF$ store their inputs as activations for gradient computation, which can lead to additional memory overhead. To mitigate this issue, we propose to store the output of normalization $\mA=\mathrm{Norm}(\mX)$ as the activation, and recover the partial derivatives as:
\begin{gather}
    \bar{\mX}=\mA/\bm{\gamma} \\
    \nabla_{\bar{\mX}}L=\nabla_{\mA}L\cdot\bm{\gamma} \\
    \nabla_{\mX}L=\frac{1}{\mathrm{RMS}}\left(\nabla_{\bar{\mX}}L-\bar{\mX}\cdot\frac{1}{n}\sum_{j=1}^{n}\left[\nabla_{\bar{\mX}}L\cdot\bar{\mX}\right]_{\cdot,j}\right)
\end{gather}
This strategy allows the $\mathrm{Norm}$ function and the sub-layer $\gF$ to share the same activation memory, since the output of $\mathrm{Norm}$ is exactly the input of $\gF$. Note that the $\mathrm{RMS}$ value should also be stored for computing $\nabla_{\mX}L$, but its size is relatively small (a one-dimensional vector) and the memory cost is negligible.
Moreover, by applying a structure-agnostic mechanism for the sub-layer $\gF$, we equip \algo\ with the flexibility to adapt to complex sub-layers, which may undergo frequent updates with the evolvement of latest foundation models.

\section{Experiments}
\subsection{Experimental setting}
To assess the generalizability of \algo, we evaluate its performance on both language and vision tasks.
For language task, we consider instruction tuning and language modeling problems. 
For instruction tuning problem, we fine-tune models on the Alpaca~\citep{taori2023stanford} and FLAN-v2~\citep{pmlr-v202-longpre23a} instruction datasets. Following prior work~\citep{dettmers2023qlora} , we subsample 50k training examples from FLAN-v2 to match the scale of Alpaca. The fine-tuned models are evaluated on the MMLU benchmark~\citep{hendrycksmeasuring} using the average 5-shot metric,
following the settings of prior works~\citep{dettmers2023qlora}.
For language modeling problem, we fine-tune model on the WikiText-2 dataset~\citep{merity2017pointer} and evaluate the performance using the perplexity metric.

For vision task, we evaluate on image classification using two datasets, including CIFAR-100~\citep{krizhevsky2009learning} and Food-101~\citep{bossard14}, and calculate the top-1 accuracy on their corresponding test datasets.

We fine-tune the LLaMa2-7B model~\citep{touvron2023llama} for language tasks. 
The learning rate is set to $3 \times 10^{-5}$ with a batch size of 128, and a micro batch size of 16. The context length is set to 256 for Alpaca and FLAN-v2, and 128 for WikiText-2.
For vision tasks, we fine-tune the ViT-B/16 model~\citep{dosovitskiy2021an} with a learning rate of 0.1, a batch size of 2048 and a micro-batch size of 1024.
For all experiments, we employ LoRA~\citep{hu2022lora} modules with a bottleneck dimension of 64 and perform fine-tuning in BF16 precision. We apply gradient checkpointing to the sub-layers of each Transformer layer to automatically compute their gradients by storing only the inputs. All experiments are conducted using a single GPU with 32GB of memory.

\begin{table}[!t]
  \setlength{\abovecaptionskip}{0.1in}
  \setlength{\belowcaptionskip}{0.0in}
  \caption{Evaluation scores and memory usage on Alpaca and FLAN-v2.}
  \label{tab:it}
  \centering
  \small
  \setlength{\tabcolsep}{0.04in} % column spacing
  \begin{tabular}{l|ccc|ccc}
    \toprule
    \multirow{2}{*}{\textbf{Methods}} & \multicolumn{3}{c|}{\textbf{Alpaca}} & \multicolumn{3}{c}{\textbf{FLAN-v2}} \\
     & \textbf{Score} ($\uparrow$) & \textbf{Total Mem.} & \textbf{Act. Mem.} & \textbf{Score} ($\uparrow$) & \textbf{Total Mem.} & \textbf{Act. Mem.}\\
    \midrule
    Base model & 45.89 & 12.465G & - & 45.89 & 12.465G & - \\
    LoRA  & 46.18 (+0.29) & 29.748G & 16.847G & 46.04 (+0.15) & 29.748G & 16.847G \\
    \algo\ ($r=1/2$) & 46.29 (+0.40) & 15.584G & 2.712G & 45.71 (-0.18) & 15.584G & 2.712G \\
    \algo\ ($r=1/4$) & 46.41 (+0.52) & 14.599G & 1.728G & 45.95 (+0.06) & 14.599G & 1.728G \\
    \algo\ ($r=1/8$) & 46.44 (+0.55) & 14.110G & 1.238G & 45.90 (+0.01) & 14.110G & 1.238G \\
    \algo\ ($r=1/16$) & 46.60 (+0.71) & 13.861G & 0.989G & 45.76 (-0.13) & 13.861G & 0.989G \\
    \algo\ ($r=1/32$) & 46.31 (+0.42) & 13.738G & 0.866G & 46.20 (+0.31) & 13.738G & 0.866G \\
    \algo\ ($r=1/64$) & 46.29 (+0.40) & 13.676G & 0.805G & 45.86 (-0.03) & 13.676G & 0.805G \\
    \bottomrule
  \end{tabular}
\end{table}

\begin{table}[!t]
  \setlength{\abovecaptionskip}{0.1in}
  \setlength{\belowcaptionskip}{0.0in}
  \caption{Perplexity and memory usage on WikiText-2.}
  \label{tab:lm}
  \centering
  \small
  \begin{tabular}{l|ccc}
    \toprule
    \textbf{Methods} & \textbf{Perplexity} ($\downarrow$) & \textbf{Total Memory} & \textbf{Activation Memory} \\
    \midrule
    Base model & 5.11 & 12.323G & - \\
    LoRA  & 4.84 (-0.27) & 19.967G & 7.566G \\
    \algo\ ($r=1/2$) & 4.93 (-0.18) & 13.441G & 1.465G \\
    \algo\ ($r=1/4$) & 4.96 (-0.15) & 13.172G & 1.056G \\
    \algo\ ($r=1/8$) & 4.98 (-0.13) & 13.047G & 0.786G \\
    \algo\ ($r=1/16$) & 5.20 (+0.09) & 12.984G & 0.662G \\
    \bottomrule
  \end{tabular}
\end{table}

\begin{table}[!t]
  \setlength{\abovecaptionskip}{0.1in}
  \setlength{\belowcaptionskip}{0.0in}
  \caption{Comparison of layer-wise and pre-norm activation compression on Alpaca and FLAN-v2.}
  \label{tab:it-compare}
  \centering
  \small
  \begin{tabular}{llcccccc}
    \toprule
    \textbf{Datasets} & \textbf{Methods} & $r=1/2$ & $r=1/4$ & $r=1/8$ & $r=1/16$ & $r=1/32$ & $r=1/64$ \\
    \midrule
    \multirow{2}{*}{Alpaca} & layer-wise & 46.24 & 46.27 & 46.30 & 46.33 & 46.25 & 46.16 \\
     & pre-norm & \textbf{46.29} & \textbf{46.41} & \textbf{46.44} & \textbf{46.60} & \textbf{46.31} & \textbf{46.29} \\
     \midrule
    \multirow{2}{*}{FLAN-v2} & layer-wise & 45.32 & 45.90 & 45.85 & 45.04 & 45.39 & 45.27 \\
     & pre-norm & \textbf{45.71} & \textbf{45.95} & \textbf{45.90} & \textbf{45.76} & \textbf{46.20} & \textbf{45.86} \\
    \bottomrule
  \end{tabular}
\end{table}

\begin{table}[!t]
  \setlength{\abovecaptionskip}{0.1in}
  \setlength{\belowcaptionskip}{0.0in}
  \caption{Comparison of layer-wise and pre-norm activation compression on WikiText-2.}
  \label{tab:lm-compare}
  \centering
  \small
  \begin{tabular}{llcccc}
    \toprule
    \textbf{Datasets} & \textbf{Methods} & $r=1/2$ & $r=1/4$ & $r=1/8$ & $r=1/16$ \\
    \midrule
    \multirow{2}{*}{WikiText-2} & layer-wise & 4.97 & 4.98 & 5.01 & 5.97 \\
     & pre-norm & \textbf{4.93} & \textbf{4.96} & \textbf{4.98} & \textbf{5.20} \\
    \bottomrule
  \end{tabular}
\end{table}

\subsection{instruction tuning and language modeling}

We first evaluate the effectiveness of \algo\ when adapted to language models. Specifically, we experiment on the instruction tuning task using Alpaca and FLAN-v2 datasets, and compare the base model, LoRA, and \algo\ under varying compression ratios ($r=k/n$). \Cref{tab:it} reports the evaluation metrics along with total and activation memory consumption. Across all settings, \algo\ can substantially reduce the memory usage while preserving the fine-tuning performance. Particularly, the activation memory falls below 1G when setting $r$ to less than $1/16$. Notebly, smaller compression ratio $r$ can even yield better performance, such as $r=1/16$ on Alpaca and $r=1/32$ on FLAN-v2. This indicates that compressing activations by discarding small singular values may even facilitate the activation quality.

Additionally, we conduct experiments on the language modeling task and report the results in \Cref{tab:lm}. Specially, we use the WikiText-2 dataset and compare the base model, LoRA and \algo\ under varying compression ratios ($r=k/n$). The results show that \algo\ attains performance comparable to LoRA fine-tuning while reduing the activation memory usage. It is worth noting that as $r $ increases, the corresponding perplexity generally decreases, which indicates that retaining more activation information benefits language modeling performance.

Moreover, we compare the proposed pre-norm activation compression with the layer-wise compression scheme, and present the results in \Cref{tab:it-compare,tab:lm-compare}. Across all settings, the pre-norm activation compression strategy consistently achieves higher score. We attribute this to greater error accumulation under layer-wise activation compression, whereas pre-norm activation compression mitigates error propagation and yields more accurate gradients during the backward pass.

\begin{table}[!t]
  \setlength{\abovecaptionskip}{0.1in}
  \setlength{\belowcaptionskip}{0.0in}
  \caption{Accuracy and memory usage on CIFAR-100 and Food-101.}
  \label{tab:cv}
  \centering
  \small
  \setlength{\tabcolsep}{0.04in} % column spacing
  \begin{tabular}{l|ccc|ccc}
    \toprule
    \multirow{2}{*}{\textbf{Methods}} & \multicolumn{3}{c|}{\textbf{CIFAR-100}} & \multicolumn{3}{c}{\textbf{Food-101}} \\
     & \textbf{Score} ($\uparrow$) & \textbf{Total Mem.} & \textbf{Act. Mem.} & \textbf{Score} ($\uparrow$) & \textbf{Total Mem.} & \textbf{Act. Mem.}\\
    \midrule
    LoRA  & 88.94 & 28.275G & 28.080G & 82.54 & 28.275G & 28.080G \\
    \algo\ ($r=1/2$) & 88.27 (-0.67) & 5.104G & 4.765G & 81.68 (-0.86)  & 5.104G & 4.765G \\
    \algo\ ($r=1/4$) & 87.10 (-1.84) & 4.230G & 3.891G & 81.52 (-1.02) & 4.230G & 3.891G \\
    \algo\ ($r=1/8$) & 81.03 (-7.91) & 3.793G & 3.454G & 78.80 (-3.74) & 3.793G & 3.454G \\
    \algo\ ($r=1/16$) & 75.97 (-12.97) & 3.575G & 3.236G & 64.16 (-18.38) & 3.575G & 3.236G \\
    \bottomrule
  \end{tabular}
\end{table}

\begin{table}[!t]
  \setlength{\abovecaptionskip}{0.1in}
  \setlength{\belowcaptionskip}{0.0in}
  \caption{Comparison of layer-wise and pre-norm activation compression on two image datasets.}
  \label{tab:cv-compare}
  \centering
  \small
  \setlength{\tabcolsep}{0.1in} % column spacing
  \begin{tabular}{lcccccccc}
    \toprule
    \multirow{2.5}{*}{\textbf{Methods}} & \multicolumn{4}{c}{\textbf{CIFAR-100}} & \multicolumn{4}{c}{\textbf{Food-101}} \\
    \cmidrule(lr){2-5}\cmidrule(lr){6-9}
    & $1/2$ & $1/4$ & $1/8$ & $1/16$ & $1/2$ & $1/4$ & $1/8$ & $1/16$ \\
    \midrule
    layer-wise & 88.01 & 84.12 & 61.19 & 7.79 & 81.05 & 80.55 & 75.42 & 4.13 \\
    pre-norm & \textbf{88.27} & \textbf{87.10} & \textbf{81.03} & \textbf{75.97} & \textbf{81.68} & \textbf{81.52} & \textbf{78.80} & \textbf{64.16} \\
    \bottomrule
  \end{tabular}
\end{table}

\subsection{Image Classification}

To assess the generalizability of \algo\ to vision tasks, we conduct experiments on image classification tasks including CIFAR-100 and Food-101, and report the results in \Cref{tab:cv}. The results show that \algo\ substantially reduces memory usage across all settings. When setting $r=1/2$ or $1/4$, \algo\ can achieve performance comparable to LoRA on both the CIFAR-100 and Food-101 datasets. Notably, when $r<1/8$, accuracy tends to decline, suggesting that, for vision tasks, \algo\ benefits from relatively larger $r$  to preserve sufficient information.

Moreover, we compare the performance of \algo\ using pre-norm activation compression against the layer-wise compression approach. The experimental results are reported in \Cref{tab:cv-compare}. Consistent with the language task, the pre-norm activation compression strategy outperforms the layer-wise compression strategy across all settings. These results further corroborate the effectiveness of the proposed pre-norm activation compression compared with the layer-wise compression strategy.

\section{Conclusion}
In this paper, we propose a novel memory-efficient fine-tuning approach \textit{Low-Rank Activation Compression} (\algo), which is motivated by the low-rank nature of activations. \algo\ introduces a flexible and versatile compression strategy that can be applied online during the forward pass, eliminating the need for calibration data. Furthermore, \algo\ incorporates a novel sampling-based orthogonal decomposition tailored to low-rank matrices, yielding greater computational efficiency and tighter error bounds. Extensive experiments demonstrate that \algo\ can achieve competitive fine-tuning performance across both language and vision tasks while reducing the activation memory usage by approximately 80\% compared to the broadly adopted LoRA method. We hope our method can assist the community in exploring the potential of memory-efficient fine-tuning.
% \subsubsection*{Author Contributions}
% If you'd like to, you may include  a section for author contributions as is done
% in many journals. This is optional and at the discretion of the authors.

% \subsubsection*{Acknowledgments}
% Use unnumbered third level headings for the acknowledgments. All
% acknowledgments, including those to funding agencies, go at the end of the paper.

\bibliography{reference}
\bibliographystyle{iclr2026_conference}

\newpage
\appendix
\section{Appendix}

\subsection{Proof of Theorem~\ref{4.0}}
We assume that in the normal case the activation matrices are denoted by $\mA_1$,$\mA_2$,…$\mA_N$, while in the forward-compression case the corresponding activation matrices are denoted by $\mA_1^{'}$, $\mA_2^{'}$,…$\mA_N^{'}$.
By the given condition, we obtain $||\mA_1-\mA_1^{'}|| \leq \sigma_1$. 

Invoking the Lipschitz continuity of $f_1$, it follows that 
$$||f_1(\mA_1)-f_1(\mA_1^{'})||\leq L||\mA_1-\mA_1^{'}||\leq L \sigma_1$$
In the normal case, we have $\mA_2=f_1(\mA_1)$. While in the forward-compression case, we further compress $f_1(\mA_1^{'})$ to obtain $\mA_2^{'}$.
Similarly, $||\mA_2^{'}-f_1(\mA_1^{'})||\leq \sigma_2$.

By the triangle inequality of the norm, we obtain:
$$
||\mA_2-\mA_2^{'}||\leq L\sigma_1+\sigma_2
$$
Proceeding in the same manner, we obtain:
$$
||\mA_N-\mA_N^{'}||\leq \Sigma_{i=1}^{N}(L^{N-i}\sigma_i)
$$
Subsequently, the final activation is passed through the model’s linear layer, which is necessarily a Lipschitz continuous function. We assume its Lipschitz constant is denoted by $L_{\mW}$.

Moreover, since the cross-entropy loss is globally Lipschitz continuous with respect to the logits, with Lipschitz constant $\sqrt{2}$.

Thus, we obtain the desired result:
$$
\mathcal{L}(F(\vx),\vy)-\mathcal{L}(F_{comp}(\vx),\vy) \leq \sqrt{2}L_{\mW}\Sigma_{i=1}^{N}(L^{N-i}\sigma_i)
$$
\qed

\subsection{Proof of Theorem \ref{4.1}}
Since $\mG$ is a standard Gaussian matrix, it is with high probability of full row rank. As our goal is to derive a lower bound for $\mathbb{E}||\frac{1}{l}\mG^T\mG\mA-\mA||$, we may assume that $\mG$ is indeed of full row rank. In this case, $\operatorname{rank}(\mG) = m - l = k$. 

We denote by $N = \operatorname{Null}(\mG)$ the null space of $\mG$, with $\dim(N) = k$. Let $\mP_{N}$ denote the orthogonal projection onto $N$. Then $\forall \vx \in \R^m$, we have:
\begin{align*}
||(\frac{1}{l}\mG^T\mG\mA-\mA)\vx||^2&=||(\frac{1}{l}\mG^T\mG-\mI)\mA\vx||^2 \\
&=||(\frac{1}{l}\mG^T\mG-\mI)\mP_N\mA\vx||^2+||(\frac{1}{l}\mG^T\mG-\mI)(\mI-\mP_N)\mA\vx||^2 \\
&\geq||(\frac{1}{l}\mG^T\mG-\mI)\mP_N\mA\vx||^2=||\mP_N\mA\vx||^2
\end{align*}
Since this inequality holds for all $\vx$, we may take the supremum over both sides to obtain:
$$
||(\frac{1}{l}\mG^T\mG\mA-\mA)|| \geq ||\mP_N\mA||
$$
Let $\vv_{1}$, $\vu_{1}$, and $\sigma_{1}$ denote the first left singular vector, the first right singular vector, and the largest singular value of $\mA$, respectively. we have:
$$
||\mP_N\mA||=\max_{||\vx||=1} ||\mP_N\mA\vx|| \geq||\mP_N\mA\vv_1||=||\mP_N\sigma_1 \vu_1||=||\mA||||\mP_N\vu_1||
$$
Therefore, $\mathbb{E}||\mP_N\mA|| \geq \mathbb{E}||\mA||||\mP_N\vu_1||=||\mA|| \mathbb{E}||\mP_N\vu_1||$. It thus remains to analyze $\mathbb{E}||\mP_N\vu_1||$.

Because $\mG$ is a standard Gaussian matrix independent of $\mA$, $N$ is a uniformly random $k$-dimensional subspace of $\R^m$
on the Grassmann manifold. For any fixed unit vector $\vu_1$, the random variable $||\mP_N\vu_1||^2$ has the same distribution as the squared length obtained by projecting a uniformly distributed vector on the unit sphere $S^{m-1}$ onto a random 
$k$-dimensional subspace. A classical result states that:
$$
||\mP_N\vu_1||^2 \sim \mathrm{Beta}(\frac{k}{2},\frac{m-k}{2})
$$
Therefore,
$\mathbb{E}||\mP_N\vu_1||=\frac{\Gamma(\frac{k+1}{2})\Gamma(\frac{m}{2})}{\Gamma(\frac{k}{2})\Gamma(\frac{m+1}{2})} \geq
\sqrt{\frac{m-l}{m+1}}$. So $\mathbb{E}||\mP_N\mA|| \geq \sqrt{\frac{m-l}{m+1}}\mathbb{E}||\mA||$. 
Hence, we have established the conclusion of Theorem~\ref{4.1}. \qed

\subsection{Proof of Theorem \ref{4.2}}
For ease of exposition, we first introduce some notation:
\begin{align*}
\mA=\mU \mSigma \mV^T
=\mU  \begin{bmatrix} \mSigma_1 & \bm0 \\ \bm0 & \mSigma_2 \\ \bm0 & \bm0  \end{bmatrix} 
\begin{bmatrix} \mV_1^T \\ \mV_2^T \end{bmatrix} 
\end{align*}

\begin{align*}
\mY=\mA \bm{\Omega}= \mQ \mR
\end{align*}
We denote that $\bm{\Omega}_1=\mV_1^T\bm{\Omega}, \bm{\Omega}_2=\mV_2^T\bm{\Omega}$, then
\begin{align*}
    \mY=\mU \mSigma \begin{bmatrix} \bm{\Omega}_1 \\ \bm{\Omega}_2 \end{bmatrix}
       =\mU \begin{bmatrix} \mSigma_1\bm{\Omega}_1 \\ \mSigma_2\bm{\Omega}_2 \\ \bm0 \end{bmatrix}
\end{align*}
\textit{Proof.} 
We now begin the proof.
We denote that $\mA_r=\mU^T\mA=\begin{bmatrix} \mSigma_1\mV_1 \\ \mSigma_2\mV_2 \\ \bm0 \end{bmatrix}$,
$\mY_r=\mA_r\bm{\Omega}=\begin{bmatrix} \mSigma_1\bm{\Omega}_1 \\ \mSigma_2\bm{\Omega}_2 \\ \bm0 \end{bmatrix}$.

Then it can be derived 
$$
||\mA -\mQ\mQ^T\mA ||=||(\mI-\mP_\mY)\mA||=||\mU^T(\mI-\mP_\mY)\mU\mA_r||=||(\mI-\mP_{\mY_r})\mA_r||
$$
where $\mP_\mY$ represents an orthogonal projection operator that satisfies $\mathrm{Range}(\mP_\mY) = \mathrm{Range}(\mY)$.

Firstly, we consider the simpler condition where $\mathrm{rank}(\mA)\leq k$. 
At this point, it is evident that $\mSigma_2 = \mathbf{0}$, thus we can observe:
$$
\mA_r=\begin{bmatrix} \mSigma_1\mV_1 \\ \bm0 \\ \bm0 \end{bmatrix} \quad \text{and} \quad  
\mY_r=\begin{bmatrix} \mSigma_1\bm{\Omega}_1 \\ \bm0 \\ \bm0 \end{bmatrix}
$$
Since $\bm{\Omega}_1$ is of full row rank, it is evident that $\mathrm{Range}(\mSigma_1\mV_1)=\mathrm{Range}(\mSigma_1\bm{\Omega}_1)$.
That is $\mathrm{Range}(\mA_r)=\mathrm{Range}(\mY_r)$. Thus, $\mathrm{Range}(\mA_r)=\mathrm{Range}(\mP_{\mY_r})$. 
As $\mP_{\mY_r}$ is the orthogonal projector onto $\mathrm{Range}(\mA_r)$, for any vector $\vx$, 
$\mP_{\mY_r}\mA_r\vx=\mA_r\vx$. Thus, we conclude that $\mP_{\mY_r}\mA_r=\mA_r$  and thereby establish the conclusion of the theorem \ref{4.2}:
$$
||\mA -\mQ\mQ^T\mA ||=||(\mI-\mP_{\mY_r})\mA_r||=0
$$
In the following, we address the more involved case where $\mathrm{rank}(\mA)>k$. 
The idea of the proof is to reformulate $\mP_{\mY_r}$  as a matrix with a more tractable structure, thereby facilitating a bound on $||\mA -\mQ\mQ^T\mA ||$. Since $\mathrm{rank}(\mA)>k$, the matrix $\mSigma_1$ is $k$-rank full, and as a result, $\mSigma_1\bm{\Omega}_1$ is of full row rank. Thus, we can reformulate $\mP_{\mY_r}$ as a block matrix,  with one block being the identity matrix $\mI_k$,while the remaining part is considered as an interference term, which allows us to perform the scaling. Specifically: let 
$$
\mZ=\mY_r\cdot\bm{\Omega}_1^{+}\mSigma_1^{-1}=\begin{bmatrix} \mI \\ \mF \\ \mathbf{0} \end{bmatrix}   
\text{, where } \mF=\mSigma_2\bm{\Omega}_2\bm{\Omega}_1^{+}\mSigma_1^{-1}
$$
It follows immediately that $\mathrm{Range}(\mZ)\subseteq \mathrm{Range}(\mY_r)$, 
Consequently, $\mathrm{Range}(\mP_{\mZ})\subseteq \mathrm{Range}(\mP_{\mY_r})$.
Hence, we derive the bound:
$$
||(\mI-\mP_{\mY_r})\mA_r|| \leq ||(\mI-\mP_{\mZ})\mA_r||
$$
We know that $||\mA||$ equals the largest singular value of $\mA$, also the square root of the largest eigenvalue of $\mA^T\mA$, 
we can further bound it as follows:
$$
||(\mI-\mP_{\mZ})\mA_r||^2=\lambda(\mA_r^T(\mI-\mP_{\mZ})^T(\mI-\mP_{\mZ})\mA_r)
=||\mA_r^T(\mI-\mP_{\mZ})\mA_r||=||\mSigma^T(\mI-\mP_{\mZ})\mSigma||
$$
The identities follow from the fact that an orthogonal projection operator is always idempotent, and the eigenvalues of a positive semi-definite matrix coincide with its singular values.

As $\mZ$ has full column rank, it follows directly from the properties of orthogonal projection operators that:
$$
\mP_{\mZ}=\mZ(\mZ^T\mZ)^{-1}\mZ^T=
\begin{bmatrix} (\mI+\mF^T\mF)^{-1} & (\mI+\mF^T\mF)^{-1}F^T & \mathbf{0} \\
                 \mF(\mI+\mF^T\mF)^{-1} & \mF(\mI+\mF^T\mF)^{-1}\mF^T & \mathbf{0}\\ 
                 \mathbf{0} &\mathbf{0} &\mathbf{0}  \end{bmatrix}
$$
Therefore, $$\mI-\mP_{\mZ}=\begin{bmatrix} \mI-(\mI+\mF^T\mF)^{-1} & -(\mI+\mF^T\mF)^{-1}\mF^T & \mathbf{0} \\
                 -\mF(\mI+\mF^T\mF)^{-1} & \mI-\mF(\mI+\mF^T\mF)^{-1}\mF^T & \mathbf{0}\\ 
                 \mathbf{0} &\mathbf{0} &\mI  \end{bmatrix}
           $$
According to \citet{DBLP:journals/siamrev/HalkoMT11}, for any positive semidefinite matrix $\mM \succeq \mathbf{0}$, we have $\mI-(\mI+\mM)^{-1}\preceq \mM$. Clearly $\mF^T\mF\succeq \mathbf{0}$,
hence, $\mI-(\mI+\mF^T\mF)^{-1}\preceq \mF^T\mF$. Moreover, since $\mI-\mF(\mI+\mF^T\mF)^{-1}\mF^T\preceq\mI$, we can obtain the following strong conclusion:
$$
\mI-\mP_{\mZ} \preceq \begin{bmatrix} \mF^T\mF & -(\mI+\mF^T\mF)^{-1}\mF^T & \mathbf{0} \\
                                       -\mF(\mI+\mF^T\mF)^{-1} & \mI & \mathbf{0}\\ 
                                       \mathbf{0} &\mathbf{0} &\mI  \end{bmatrix}
$$
Thus, 
$$
\mSigma^T(\mI-\mP_{\mZ})\mSigma \preceq \begin{bmatrix} \mSigma_1\mF^T\mF\mSigma_1 &  -\mSigma_1(\mI+\mF^T\mF)^{-1}\mF^T\mSigma_2\\
                                       -\mSigma_2\mF(\mI+\mF^T\mF)^{-1}\mSigma_1 & \mSigma_2^2\\ 
                                         \end{bmatrix}
$$
Note that $(-\mSigma_1(\mI+\mF^T\mF)^{-1}\mF^T\mSigma_2)^T= -\mSigma_2\mF(\mI+\mF^T\mF)^{-1}\mSigma_1$, 
therefore, according to  \citet{DBLP:journals/siamrev/HalkoMT11} we have
$$
||\mSigma^T(\mI-\mP_{\mZ})\mSigma|| \leq ||\mSigma_1\mF^T\mF\mSigma_1||+||\mSigma_2^2||=||\mSigma_2||^2+||\mF\mSigma_1||^2=
||\mSigma_2||^2+||\mSigma_2\bm{\Omega}_2\bm{\Omega}_1^+||^2
$$
Combining the above arguments, we complete the proof. \qed

\subsection{Proof of Theorem \ref{4.3}} 
Our method is similar to RSVD in that it employs a sampling matrix $\mathbf{\Omega}$ to sample the row space of $\mA$, 
approximately capturing the top-$k$ left singular vectors of $\mA$.  
Consequently, the result of Theorem \ref{4.2} plays a crucial role, motivating our effort to revise and present a corrected proof of it.

Theorem \ref{4.2} states that when $\mathbf{\Omega_1}$ has full column rank, the approximation error admits a deterministic upper bound:
$$|| \mA -\mQ\mQ^T\mA ||^2 \leq ||\mSigma_2||^2+||\mSigma_2\bm{\Omega}_2\bm{\Omega}_1^+||^2$$
By a simple relaxation we further obtain:
$$
|| \mA -\mQ\mQ^T\mA || \leq ||\mSigma_2|| + ||\mSigma_2\bm{\Omega}_2\bm{\Omega}_1^+||
$$

In the classical RSVD method, the sampling matrix $\mathbf{\Omega}$ is taken to be a standard Gaussian matrix, 
so $\mathbf{\Omega}_1$ is almost surely of full row rank, the upper bound for RSVD can be directly derived by invoking the conclusion of Theorem \ref{4.2} and applying suitable relaxations based on the properties of standard Gaussian matrices. 
However, in our method, $\mathbf{\Omega}$ relies on uniform random sampling from $\mA$, 
and thus $\mathbf{\Omega}_1$ is not guaranteed to be of full row rank under all realizations.

Our proof strategy is to first analyze the quantities $||\mSigma_2\bm{\Omega}_2\bm{\Omega}_1^+||$ and $\mathbf{\Omega}_1$, and then define a favorable event accordingly. 
This favorable event occurs with high probability and simultaneously satisfies Theorem \ref{4.2}, while its complement occurs with low probability. 
We employ Theorem \ref{4.2} to bound the expectation of $||\mA - \mQ\mQ^T\mA||$ when restricted to the favorable event, and use probability arguments to control the expectation restricted to the complement of the favorable event. In this way, we obtain an upper bound on $\mathbb{E}||\mA - \mQ\mQ^T\mA||$.

First, we analyze the quantities $||\mSigma_2\bm{\Omega}_2\bm{\Omega}_1^+||$ and $\bm{\Omega}_1$. 
We define the sampling matrix as $\mR_{l \times m}$, then 
$$\mathbf{\Omega}=(\mR\mA)^T=\mA^T\mR^T=(\mU\mSigma\mV)^T\mR^T=\mV\mSigma^T\mU^T\mS^T$$
Therefore, according to the previous definitions, we have $\mathbf{\Omega}_1=[\mSigma_1,\mathbf{0}]\mU^T\mR^T$, $\mathbf{\Omega}_2=[\mathbf{0},\mSigma_2,\mathbf{0}]\mU^T\mR^T$.
Thus,
$$
||\mSigma_2\bm{\Omega}_2\bm{\Omega}_1^+||=||\mSigma_2 [\mathbf{0},\mSigma_2,\mathbf{0}]\mU^T\mR^T ([\mSigma_1,\mathbf{0}]\mU^T\mR^T)^+||
$$
In order to extract $\mSigma_1^+$ from $([\mSigma_1,\mathbf{0}] \mU^T \mR^T)^+$, we reformulate $||\mSigma_2 \bm{\Omega}_2 \bm{\Omega}_1^+||$ as follows:
$$
||\mSigma_2\mathbf{\Omega}_2\mathbf{\Omega}_1^+||=||\mSigma_2(\mSigma_2\mU_{n-k}^T\mR^T)(\mSigma_1\mU_k^T\mR^T)^+||=
||\mSigma_2(\mSigma_2\mU_{n-k}^T\mR^T)(\mU_k^T\mR^T)^+\mSigma_1^+||
$$
where $\mU_k$ denotes the first $k$ rows of $\mU$, and $\mU_{n-k}$ denotes the $(k+1)$-th through $n$-th rows of $\mU$.
By a simple rescaling, we obtain
\begin{align*}
   ||\mSigma_2\mathbf{\Omega}_2\mathbf{\Omega}_1^+|| 
   &\leq ||\mSigma_2|| ||(\mSigma_2\mU_{n-k}^T\mR^T)(\mU_k^T\mR^T)^+|| ||\mSigma_1^+|| \\
   &=\frac{\sigma_{k+1}(A)}{\sigma_{k}(A)}||(\mSigma_2\mU_{n-k}^T\mR^T)(\mU_k^T\mR^T)^+||\\                                       
   &\leq ||\mSigma_2(\mU_{n-k}^T\mR^T)(\mU_k^T\mR^T)^+||
\end{align*}

We now define the good event.  For any $\epsilon\in(0,1/2)$, we define
$$\mathcal{G}(\epsilon)=\{\sigma_{min}(\mU_k^T)\geq\sqrt{1-\epsilon} \land ||\mU_{n-k}^T\mR^T||\leq 1+\epsilon \}$$

Recall the notion of coherence: for an orthonormal matrix $\mU$, its coherence is defined as
$\mu_k:=\frac{m}{k}\,\max_{1 \le i \le m} || (\mU_k)_{i,:} ||_2$.

According to the classical matrix Chernoff–type results~\citep{DBLP:journals/ftml/Tropp15,DBLP:conf/nips/CohenPW20}, 
for every $\delta\in(0,1)$,  if $ l\geq C\frac{\mu_k k}{\epsilon^2}\log{\frac{k}{\delta}}$,
then $\mathbb{P}(\mathcal{G}(\epsilon))\geq  1- \delta$.

Based on the subspace principal angle lemma and the conclusions of previous work~\citep{DBLP:journals/corr/abs-1008-0587,DBLP:conf/icalp/CohenNW16}, we can further deduce that:
$$
\mathbb{E}||(\mU_{n-k}^T\mR^T)(\mU_k^T\mR^T)^+|| \leq C\sqrt{\frac{\mu_k k}{l}}
$$
Thus, $\mathbb{E}||\mSigma_2\mathbf{\Omega}_2\mathbf{\Omega}_1^+|| \leq (C\sqrt{\frac{\mu_k k}{l}}) \sigma_{k+1}(\mA)$.
$\mathbb{E}||\mA - \mQ\mQ^T\mA||\leq (1+C\sqrt{\frac{\mu_k k}{l}}) \sigma_{k+1}(\mA)$, which is the expectation of $||\mA - \mQ\mQ^T\mA||$ when restricted to the favorable event.

Finally, we derive the expectation restricted to the complement of the favorable event. It is evident:
$$
\mathbb{E}||\mA - \mQ\mQ^T\mA|| =\mathbb{E}||(\mI-\mP_{\mY})\mA|| \leq \mathbb{E}||\mA||= ||\mA|| \mathbb{P}(\mathcal{G}^c)
\leq \delta||\mA|| \leq ke^{\frac{-l\epsilon^2}{C\mu_k k}}||\mA||.
$$
Since $\epsilon$ is an arbitrary positive real number, we may set $\epsilon=\sqrt{\frac{\mu_k k}{l}}$. Substituting this choice yields:
$$
\delta||\mA|| \leq ke^{\frac{1}{C}} ||\mA||
$$
By adding the expectation of $||\mA - \mQ\mQ^T\mA||$ restricted to the favorable event and the expectation restricted to its complement, we obtain the desired conclusion. \qed

\subsection{Related Work}

\paragraph{Parameter-Efficient Fine-Tuning}
% Parameter-efficient fine-tuning (PEFT) diminishes the number of trainable parameters, thereby lessening the memory demands to some extent.
Parameter-efficient fine-tuning (PEFT) provides an effective strategy for adapting pre-trained models to downstream tasks. By modifying only a small subset of parameters, these methods achieve competitive performance compared with full fine-tuning.
The additive PEFT methods, like Adapter Tuning~\citep{houlsby2019parameter}, Prompt Tuning~\citep{lester2021power} and Prefix Tuning~\citep{li2021prefix}, keep the original model weights frozen and introduce new trainable components, but these approaches increase inference latency. Reparameterization-based PEFT methods, such as Low-Rank Adaptation (LoRA)~\citep{hu2022lora}, operates by constraining trainable parameters within a low-rank subspace for each weight matrix. Besides, SPDF~\citep{thangarasa2023spdf}, S$^2$FT~\citep{yang2024s}, and SMT~\citep{he2025smt} leverage sparsity by selectively training only a subset of the model weights. However, a significant memory bottleneck still remains for existing PEFT methods, as these methods do not reduce the memory required for storing activations, thereby limiting their practicability in resource-constrained scenarios.

% \paragraph{Model Quantization}
\paragraph{Parameter Quantization and Decomposition}
The growing demand for computational efficiency highlights the deployment challenges posed by the large size of modern foundation models, particularly on edge devices.
Model quantization and decomposition serve as the principal strategies to mitigate this challenge. Existing quantization techniques generally fall into two categories: outlier-aware weight-only quantization~\citep{lin2024awq, lee2024owq} and joint activation-weight quantization~\citep{li2024svdquant, hu2025quad}. Regarding model decomposition,  a straightforward application of SVD often harms performance, since weight matrices tend to be high-rank~\citep{yu2023compressing}. To overcome this, FWSVD~\citep{hsu2022language}, ASVD~\citep{yuan2023asvd}, and WeLore ~\citep{ jaiswallow} decompose weights into low-rank matrices by preserving output similarity based on activations. LoSparse~\citep{li2023losparse} aim to recover high-rank information by additionally augmenting a low-rank component with a sparse matrix. The further approaches combine decomposition with quantization, representing weights as the sum of a low-rank component and a quantized matrix~\citep{guo2023lq}. Besides these inference-time optimizations, QLoRA~\citep{dettmers2023qlora} and LQ-LoRA~\citep{guo2023lq} reduce memory consumption for fine-tuning by first quantizing the pre-trained weights and then applying low-rank adaptation. However, the quantization and decomposition methods inevitably lead to a loss in model performance.

\paragraph{Gradient and Optimizer State Compression}
To reduce the memory consumption during training processes, recent work focuses on compressing gradients and optimizer states, considering they are naturally low-rank during training, which has been studied in both theory and practice~\citep{cosson2023low, shen2025mlorc}. Among existing approaches, GaLore~\citep{zhao2024galore} and FLoRA~\citep{DBLP:conf/icml/HaoCM24} project gradients into a low-rank subspace for optimizer updates, thereby saving the memory cost. However, GaLore relies on SVD for constructing the projection matrix, which introduces non=negligible computational overhead. Moreover, the low-rank updating subspace is discontinuous. Subsequent research aims to mitigate these issues by improving the projection matrix~\citep{muhamed2024grass, zhang2025i3s, zhao2025separate, xiao2025coap, liang2024memory} or integrating GaLore with other optimizer-focused techniques~\citep{huang2024galore, zhu2024apollo}. Different from the projection-based approaches, LoGE~\citep{zhang2024efficient} uses a low-rank decomposition of weights specifically during the backward pass to compute activation gradients, thus accelerating the fine-tuning efficiency. Nevertheless, the memory consumed by activations is often neglected. This is especially pronounced under the PEFT setting, where activations become the primary memory bottleneck. Therefore, existing strategies focusing solely on gradients and optimizer states are unsuitable for PEFT.

\subsection{Limitations and Discussions}
While \algo\ demonstrates significant memory reduction and competitive performance across a wide range of tasks, its effectiveness is sensitive to the chosen compression ratio $r$. In particular, under aggressive compression settings (e.g., $r<1/8$ for vision tasks), a noticeable decline in accuracy is observed. This suggests that the activation compression may discard some critical information under certain fine-tuning scenarios. This limitation may pose challenges for \algo\ in some complex reasoning benchmarks, where the estimation errors may accumulate and then degrade model performance. Therefore, it is crucial to explore adaptive compression strategies with optimal $r$ based on critical information, such as activation characteristics or task-specific requirements, and we leave this problem for future research.

\end{document}

%% file: math_commands.tex
\usepackage{amsmath,amsfonts,bm}

\def\eqref#1{equation~\ref{#1}}
\def\1{\bm{1}}

\def\vtheta{{\bm{\theta}}}
\def\va{{\bm{a}}}

\def\vg{{\bm{g}}}

\def\vu{{\bm{u}}}
\def\vv{{\bm{v}}}

\def\vx{{\bm{x}}}
\def\vy{{\bm{y}}}

\def\mA{{\bm{A}}}

\def\mF{{\bm{F}}}
\def\mG{{\bm{G}}}

\def\mI{{\bm{I}}}

\def\mK{{\bm{K}}}

\def\mM{{\bm{M}}}

\def\mP{{\bm{P}}}
\def\mQ{{\bm{Q}}}
\def\mR{{\bm{R}}}
\def\mS{{\bm{S}}}

\def\mU{{\bm{U}}}
\def\mV{{\bm{V}}}
\def\mW{{\bm{W}}}
\def\mX{{\bm{X}}}
\def\mY{{\bm{Y}}}
\def\mZ{{\bm{Z}}}

\def\mSigma{{\bm{\Sigma}}}

\DeclareMathAlphabet{\mathsfit}{\encodingdefault}{\sfdefault}{m}{sl}
\SetMathAlphabet{\mathsfit}{bold}{\encodingdefault}{\sfdefault}{bx}{n}

\def\gF{{\mathcal{F}}}

\def\gO{{\mathcal{O}}}

\def\sR{{\mathbb{R}}}

\newcommand{\Ls}{\mathcal{L}}
\newcommand{\R}{\mathbb{R}}

%% file: arxiv.bbl
\begin{thebibliography}{62}
\providecommand{\natexlab}[1]{#1}
\providecommand{\url}[1]{\texttt{#1}}
\expandafter\ifx\csname urlstyle\endcsname\relax
  \providecommand{\doi}[1]{doi: #1}\else
  \providecommand{\doi}{doi: \begingroup \urlstyle{rm}\Url}\fi

\bibitem[Bahdanau et~al.(2015)Bahdanau, Cho, and Bengio]{bahdanau2015neural}
Dzmitry Bahdanau, Kyunghyun Cho, and Yoshua Bengio.
\newblock Neural machine translation by jointly learning to align and translate.
\newblock In \emph{The 3rd International Conference on Learning Representations}, 2015.

\bibitem[Bossard et~al.(2014)Bossard, Guillaumin, and Van~Gool]{bossard14}
Lukas Bossard, Matthieu Guillaumin, and Luc Van~Gool.
\newblock Food-101 -- mining discriminative components with random forests.
\newblock In \emph{European Conference on Computer Vision}, 2014.

\bibitem[Cohen et~al.(2020)Cohen, Pagh, and Woodruff]{DBLP:conf/nips/CohenPW20}
Edith Cohen, Rasmus Pagh, and David Woodruff.
\newblock Wor and $p$'s: Sketches for $\ell_p$-sampling without replacement.
\newblock In \emph{Advances in Neural Information Processing Systems}, pp.\  21092--21104, 2020.

\bibitem[Cohen et~al.(2016)Cohen, Nelson, and Woodruff]{DBLP:conf/icalp/CohenNW16}
Michael~B. Cohen, Jelani Nelson, and David~P. Woodruff.
\newblock Optimal approximate matrix product in terms of stable rank.
\newblock In \emph{43rd International Colloquium on Automata, Languages, and Programming}, pp.\  11:1--11:14, 2016.

\bibitem[Cosson et~al.(2023)Cosson, Jadbabaie, Makur, Reisizadeh, and Shah]{cosson2023low}
Romain Cosson, Ali Jadbabaie, Anuran Makur, Amirhossein Reisizadeh, and Devavrat Shah.
\newblock Low-rank gradient descent.
\newblock \emph{IEEE Open Journal of Control Systems}, 2:\penalty0 380--395, 2023.

\bibitem[Dettmers et~al.(2023)Dettmers, Pagnoni, Holtzman, and Zettlemoyer]{dettmers2023qlora}
Tim Dettmers, Artidoro Pagnoni, Ari Holtzman, and Luke Zettlemoyer.
\newblock Qlora: Efficient finetuning of quantized llms.
\newblock \emph{Advances in neural information processing systems}, 36:\penalty0 10088--10115, 2023.

\bibitem[Ding et~al.(2023)Ding, Qin, Yang, Wei, Yang, Su, Hu, Chen, Chan, Chen, et~al.]{ding2023parameter}
Ning Ding, Yujia Qin, Guang Yang, Fuchao Wei, Zonghan Yang, Yusheng Su, Shengding Hu, Yulin Chen, Chi-Min Chan, Weize Chen, et~al.
\newblock Parameter-efficient fine-tuning of large-scale pre-trained language models.
\newblock \emph{Nature Machine Intelligence}, 5\penalty0 (3):\penalty0 220--235, 2023.

\bibitem[Dosovitskiy et~al.(2021)Dosovitskiy, Beyer, Kolesnikov, Weissenborn, Zhai, Unterthiner, Dehghani, Minderer, Heigold, Gelly, Uszkoreit, and Houlsby]{dosovitskiy2021an}
Alexey Dosovitskiy, Lucas Beyer, Alexander Kolesnikov, Dirk Weissenborn, Xiaohua Zhai, Thomas Unterthiner, Mostafa Dehghani, Matthias Minderer, Georg Heigold, Sylvain Gelly, Jakob Uszkoreit, and Neil Houlsby.
\newblock An image is worth 16x16 words: Transformers for image recognition at scale.
\newblock In \emph{The 9th International Conference on Learning Representations}, 2021.

\bibitem[Drineas \& Mahoney(2005)Drineas and Mahoney]{JMLR:v6:drineas05a}
Petros Drineas and Michael~W. Mahoney.
\newblock On the nystrom method for approximating a gram matrix for improved kernel-based learning.
\newblock \emph{Journal of Machine Learning Research}, 6\penalty0 (72):\penalty0 2153--2175, 2005.

\bibitem[Eckart \& Young(1936)Eckart and Young]{eckart1936approximation}
Carl Eckart and Gale Young.
\newblock The approximation of one matrix by another of lower rank.
\newblock \emph{Psychometrika}, 1\penalty0 (3):\penalty0 211--218, 1936.

\bibitem[Guo et~al.(2023)Guo, Greengard, Xing, and Kim]{guo2023lq}
Han Guo, Philip Greengard, Eric~P Xing, and Yoon Kim.
\newblock Lq-lora: Low-rank plus quantized matrix decomposition for efficient language model finetuning.
\newblock \emph{arXiv preprint arXiv:2311.12023}, 2023.

\bibitem[Halko et~al.(2011)Halko, Martinsson, and Tropp]{DBLP:journals/siamrev/HalkoMT11}
Nathan Halko, Per-Gunnar Martinsson, and Joel~A. Tropp.
\newblock Finding structure with randomness: Probabilistic algorithms for constructing approximate matrix decompositions.
\newblock \emph{SIAM Review}, 53\penalty0 (2):\penalty0 217--288, 2011.

\bibitem[Han et~al.(2024)Han, Gao, Liu, Zhang, and Zhang]{han2024parameter}
Zeyu Han, Chao Gao, Jinyang Liu, Jeff Zhang, and Sai~Qian Zhang.
\newblock Parameter-efficient fine-tuning for large models: A comprehensive survey.
\newblock \emph{Transactions on Machine Learning Research}, 2024.

\bibitem[Hao et~al.(2024)Hao, Cao, and Mou]{DBLP:conf/icml/HaoCM24}
Yongchang Hao, Yanshuai Cao, and Lili Mou.
\newblock Flora: Low-rank adapters are secretly gradient compressors.
\newblock In \emph{Proceedings of the 41st International Conference on Machine Learning}, pp.\  17554--17571, 2024.

\bibitem[He et~al.(2025)He, Li, Jiang, and Miller]{he2025smt}
Haoze He, Juncheng~B Li, Xuan Jiang, and Heather Miller.
\newblock Smt: Fine-tuning large language models with sparse matrices.
\newblock In \emph{The 13th International Conference on Learning Representations}, 2025.

\bibitem[Hendrycks et~al.(2021)Hendrycks, Burns, Basart, Zou, Mazeika, Song, and Steinhardt]{hendrycksmeasuring}
Dan Hendrycks, Collin Burns, Steven Basart, Andy Zou, Mantas Mazeika, Dawn Song, and Jacob Steinhardt.
\newblock Measuring massive multitask language understanding.
\newblock In \emph{The 9th International Conference on Learning Representations}, 2021.

\bibitem[Houlsby et~al.(2019)Houlsby, Giurgiu, Jastrzebski, Morrone, De~Laroussilhe, Gesmundo, Attariyan, and Gelly]{houlsby2019parameter}
Neil Houlsby, Andrei Giurgiu, Stanislaw Jastrzebski, Bruna Morrone, Quentin De~Laroussilhe, Andrea Gesmundo, Mona Attariyan, and Sylvain Gelly.
\newblock Parameter-efficient transfer learning for nlp.
\newblock In \emph{Proceedings of the 36th International Conference on Machine Learning}, pp.\  2790--2799, 2019.

\bibitem[Hsu et~al.(2022)Hsu, Hua, Chang, Lou, Shen, and Jin]{hsu2022language}
Yen{-}Chang Hsu, Ting Hua, Sungen Chang, Qian Lou, Yilin Shen, and Hongxia Jin.
\newblock Language model compression with weighted low-rank factorization.
\newblock In \emph{The 10th International Conference on Learning Representations}, 2022.

\bibitem[Hu et~al.(2022)Hu, Shen, Wallis, Allen-Zhu, Li, Wang, Wang, and Chen]{hu2022lora}
Edward~J. Hu, Yelong Shen, Phillip Wallis, Zeyuan Allen-Zhu, Yuanzhi Li, Shean Wang, Lu~Wang, and Weizhu Chen.
\newblock Lora: Low-rank adaptation of large language models.
\newblock In \emph{The 10th International Conference on Learning Representations}, 2022.

\bibitem[Hu et~al.(2025)Hu, Chen, Li, Chen, and Zhang]{hu2025quad}
Yuxuan Hu, Xiaodong Chen, Cuiping Li, Hong Chen, and Jing Zhang.
\newblock Quad: Quantization and parameter-efficient tuning of llm with activation decomposition.
\newblock \emph{arXiv preprint arXiv:2503.19353}, 2025.

\bibitem[Huang et~al.(2024)Huang, Zhang, Zhang, Luo, Sun, and Wang]{huang2024galore}
Weihao Huang, Zhenyu Zhang, Yushun Zhang, Zhi-Quan Luo, Ruoyu Sun, and Zhangyang Wang.
\newblock Galore-mini: Low rank gradient learning with fewer learning rates.
\newblock In \emph{NeurIPS 2024 Workshop on Fine-Tuning in Modern Machine Learning: Principles and Scalability}, 2024.

\bibitem[Huang et~al.(2025)Huang, Huang, and Wen]{huang2025sola}
Xinhao Huang, You-Liang Huang, and Zeyi Wen.
\newblock Sola: Leveraging soft activation sparsity and low-rank decomposition for large language model compression.
\newblock In \emph{Proceedings of the 39th AAAI Conference on Artificial Intelligence}, pp.\  17494--17502, 2025.

\bibitem[Jaiswal et~al.(2025)Jaiswal, Wang, Yin, Liu, Chen, Zhao, Grama, Tian, and Wang]{jaiswallow}
Ajay~Kumar Jaiswal, Yifan Wang, Lu~Yin, Shiwei Liu, Runjin Chen, Jiawei Zhao, Ananth Grama, Yuandong Tian, and Zhangyang Wang.
\newblock From low rank gradient subspace stabilization to low-rank weights: Observations, theories, and applications.
\newblock In \emph{Proceedings of the 42nd International Conference on Machine Learning}, 2025.

\bibitem[Kingma \& Ba(2015)Kingma and Ba]{kingma2014adam}
Diederik~P. Kingma and Jimmy Ba.
\newblock Adam: A method for stochastic optimization.
\newblock In \emph{The 3rd International Conference on Learning Representations}, 2015.

\bibitem[Krizhevsky et~al.(2009)Krizhevsky, Hinton, et~al.]{krizhevsky2009learning}
Alex Krizhevsky, Geoffrey Hinton, et~al.
\newblock Learning multiple layers of features from tiny images.
\newblock \emph{Master's thesis, University of Tront}, 2009.

\bibitem[Lee et~al.(2024)Lee, Jin, Kim, Kim, and Park]{lee2024owq}
Changhun Lee, Jungyu Jin, Taesu Kim, Hyungjun Kim, and Eunhyeok Park.
\newblock Owq: Outlier-aware weight quantization for efficient fine-tuning and inference of large language models.
\newblock In \emph{Proceedings of the 38th AAAI Conference on Artificial Intelligence}, pp.\  13355--13364, 2024.

\bibitem[Lester et~al.(2021)Lester, Al-Rfou, and Constant]{lester2021power}
Brian Lester, Rami Al-Rfou, and Noah Constant.
\newblock The power of scale for parameter-efficient prompt tuning.
\newblock \emph{arXiv preprint arXiv:2104.08691}, 2021.

\bibitem[Li et~al.(2024)Li, Lin, Zhang, Cai, Li, Guo, Xie, Meng, Zhu, and Han]{li2024svdquant}
Muyang Li, Yujun Lin, Zhekai Zhang, Tianle Cai, Xiuyu Li, Junxian Guo, Enze Xie, Chenlin Meng, Jun-Yan Zhu, and Song Han.
\newblock Svdquant: Absorbing outliers by low-rank components for 4-bit diffusion models.
\newblock \emph{arXiv preprint arXiv:2411.05007}, 2024.

\bibitem[Li \& Liang(2021)Li and Liang]{li2021prefix}
Xiang~Lisa Li and Percy Liang.
\newblock Prefix-tuning: Optimizing continuous prompts for generation.
\newblock \emph{arXiv preprint arXiv:2101.00190}, 2021.

\bibitem[Li et~al.(2023)Li, Yu, Zhang, Liang, He, Chen, and Zhao]{li2023losparse}
Yixiao Li, Yifan Yu, Qingru Zhang, Chen Liang, Pengcheng He, Weizhu Chen, and Tuo Zhao.
\newblock Losparse: Structured compression of large language models based on low-rank and sparse approximation.
\newblock In \emph{Proceedings of the 40th International Conference on Machine Learning}, pp.\  20336--20350, 2023.

\bibitem[Liang et~al.(2024)Liang, Liu, Chen, and Liu]{liang2024memory}
Kaizhao Liang, Bo~Liu, Lizhang Chen, and Qiang Liu.
\newblock Memory-efficient llm training with online subspace descent.
\newblock In \emph{Advances in Neural Information Processing Systems}, volume~37, pp.\  64412--64432, 2024.

\bibitem[Lin et~al.(2024)Lin, Tang, Tang, Yang, Chen, Wang, Xiao, Dang, Gan, and Han]{lin2024awq}
Ji~Lin, Jiaming Tang, Haotian Tang, Shang Yang, Wei-Ming Chen, Wei-Chen Wang, Guangxuan Xiao, Xingyu Dang, Chuang Gan, and Song Han.
\newblock Awq: Activation-aware weight quantization for on-device llm compression and acceleration.
\newblock \emph{Proceedings of machine learning and systems}, pp.\  87--100, 2024.

\bibitem[Liu et~al.(2024)Liu, Feng, Xue, Wang, Wu, Lu, Zhao, Deng, Zhang, Ruan, et~al.]{liu2024deepseek}
Aixin Liu, Bei Feng, Bing Xue, Bingxuan Wang, Bochao Wu, Chengda Lu, Chenggang Zhao, Chengqi Deng, Chenyu Zhang, Chong Ruan, et~al.
\newblock Deepseek-v3 technical report.
\newblock \emph{arXiv preprint arXiv:2412.19437}, 2024.

\bibitem[Liu et~al.(2025)Liu, Zhang, Wang, Yang, Hovland, Nicolae, Cappello, and Zhang]{liu2025cola}
Ziyue Liu, Ruijie Zhang, Zhengyang Wang, Zi~Yang, Paul Hovland, Bogdan Nicolae, Franck Cappello, and Zheng Zhang.
\newblock Cola: Compute-efficient pre-training of llms via low-rank activation.
\newblock \emph{arXiv preprint arXiv:2502.10940}, 2025.

\bibitem[Longpre et~al.(2023)Longpre, Hou, Vu, Webson, Chung, Tay, Zhou, Le, Zoph, Wei, and Roberts]{pmlr-v202-longpre23a}
Shayne Longpre, Le~Hou, Tu~Vu, Albert Webson, Hyung~Won Chung, Yi~Tay, Denny Zhou, Quoc~V Le, Barret Zoph, Jason Wei, and Adam Roberts.
\newblock The flan collection: Designing data and methods for effective instruction tuning.
\newblock In \emph{Proceedings of the 40th International Conference on Machine Learning}, pp.\  22631--22648, 2023.

\bibitem[Magdon-Ismail(2010)]{DBLP:journals/corr/abs-1008-0587}
Malik Magdon-Ismail.
\newblock Row sampling for matrix algorithms via a non-commutative bernstein bound.
\newblock \emph{arXiv preprint arXiv:1008.0587}, 2010.

\bibitem[Merity et~al.(2017)Merity, Xiong, Bradbury, and Socher]{merity2017pointer}
Stephen Merity, Caiming Xiong, James Bradbury, and Richard Socher.
\newblock Pointer sentinel mixture models.
\newblock In \emph{The 5th International Conference on Learning Representations}, 2017.

\bibitem[Miles et~al.(2024)Miles, Reddy, Elezi, and Deng]{miles2024velora}
Roy Miles, Pradyumna Reddy, Ismail Elezi, and Jiankang Deng.
\newblock Velora: Memory efficient training using rank-1 sub-token projections.
\newblock \emph{Advances in Neural Information Processing Systems}, 37:\penalty0 42292--42310, 2024.

\bibitem[Muhamed et~al.(2024)Muhamed, Li, Woodruff, Diab, and Smith]{muhamed2024grass}
Aashiq Muhamed, Oscar Li, David~P. Woodruff, Mona~T. Diab, and Virginia Smith.
\newblock Grass: Compute efficient low-memory llm training with structured sparse gradients.
\newblock In \emph{Proceedings of the Conference on Empirical Methods in Natural Language Processing}, pp.\  14978--15003, 2024.

\bibitem[Rumelhart et~al.(1986)Rumelhart, Hinton, and Williams]{rumelhart1986learning}
David~E. Rumelhart, Geoffrey~E. Hinton, and Ronald~J. Williams.
\newblock Learning representations by back-propagating errors.
\newblock \emph{Nature}, 323\penalty0 (6088):\penalty0 533--536, 1986.

\bibitem[Sakr \& Khailany(2024)Sakr and Khailany]{sakr2024espace}
Charbel Sakr and Brucek Khailany.
\newblock Espace: Dimensionality reduction of activations for model compression.
\newblock In \emph{Advances in Neural Information Processing Systems}, volume~37, pp.\  17489--17517, 2024.

\bibitem[Shamshoum et~al.(2025)Shamshoum, Hodos, Sieradzki, and Schuster]{shamshoum2024compact}
Yara Shamshoum, Nitzan Hodos, Yuval Sieradzki, and Assaf Schuster.
\newblock Compact: Compressed activations for memory-efficient llm training.
\newblock In \emph{Proceedings of Conference of the Nations of the Americas Chapter of the Association for Computational Linguistics: Human Language Technologies}, pp.\  1511--1524, 2025.

\bibitem[Shen et~al.(2025)Shen, Yaxiang, Huang, Xu, Zhang, and Shen]{shen2025mlorc}
Wei Shen, Zhang Yaxiang, Minhui Huang, Mengfan Xu, Jiawei Zhang, and Cong Shen.
\newblock Mlorc: Momentum low-rank compression for large language model adaptation.
\newblock \emph{arXiv preprint arXiv:2506.01897}, 2025.

\bibitem[Simoulin et~al.(2025)Simoulin, Park, Liu, and Yang]{simoulin2025memory}
Antoine Simoulin, Namyong Park, Xiaoyi Liu, and Grey Yang.
\newblock Memory-efficient fine-tuning of transformers via token selection.
\newblock \emph{arXiv preprint arXiv:2501.18824}, 2025.

\bibitem[Sutskever et~al.(2013)Sutskever, Martens, Dahl, and Hinton]{sutskever2013importance}
Ilya Sutskever, James Martens, George Dahl, and Geoffrey Hinton.
\newblock On the importance of initialization and momentum in deep learning.
\newblock In \emph{Proceedings of the 30th International Conference on Machine Learning}, pp.\  1139--1147, 2013.

\bibitem[Taori et~al.(2023)Taori, Gulrajani, Zhang, Dubois, Li, Guestrin, Liang, and Hashimoto]{taori2023stanford}
Rohan Taori, Ishaan Gulrajani, Tianyi Zhang, Yann Dubois, Xuechen Li, Carlos Guestrin, Percy Liang, and Tatsunori~B. Hashimoto.
\newblock Stanford alpaca: An instruction-following llama model, 2023.

\bibitem[Thangarasa et~al.(2023)Thangarasa, Gupta, Marshall, Li, Leong, DeCoste, Lie, and Saxena]{thangarasa2023spdf}
Vithursan Thangarasa, Abhay Gupta, William Marshall, Tianda Li, Kevin Leong, Dennis DeCoste, Sean Lie, and Shreyas Saxena.
\newblock Spdf: Sparse pre-training and dense fine-tuning for large language models.
\newblock In \emph{Proceedings of the 39th Conference on Uncertainty in Artificial Intelligence}, pp.\  2134--2146, 2023.

\bibitem[Touvron et~al.(2023)Touvron, Martin, Stone, Albert, Almahairi, Babaei, Bashlykov, Batra, Bhargava, Bhosale, et~al.]{touvron2023llama}
Hugo Touvron, Louis Martin, Kevin Stone, Peter Albert, Amjad Almahairi, Yasmine Babaei, Nikolay Bashlykov, Soumya Batra, Prajjwal Bhargava, Shruti Bhosale, et~al.
\newblock Llama 2: Open foundation and fine-tuned chat models.
\newblock \emph{arXiv preprint arXiv:2307.09288}, 2023.

\bibitem[Tropp et~al.(2015)]{DBLP:journals/ftml/Tropp15}
Joel~A. Tropp et~al.
\newblock An introduction to matrix concentration inequalities.
\newblock \emph{Foundations and Trends in Machine Learning}, 8\penalty0 (1-2):\penalty0 1--230, 2015.

\bibitem[Vaswani et~al.(2017)Vaswani, Shazeer, Parmar, Uszkoreit, Jones, Gomez, Kaiser, and Polosukhin]{vaswani2017attention}
Ashish Vaswani, Noam Shazeer, Niki Parmar, Jakob Uszkoreit, Llion Jones, Aidan~N Gomez, {\L}ukasz Kaiser, and Illia Polosukhin.
\newblock Attention is all you need.
\newblock In \emph{Advances in Neural Information Processing Systems}, volume~30, pp.\  5998--6008, 2017.

\bibitem[Wang et~al.(2019)Wang, Li, Xiao, Zhu, Li, Wong, and Chao]{wang2019learning}
Qiang Wang, Bei Li, Tong Xiao, Jingbo Zhu, Changliang Li, Derek~F Wong, and Lidia~S Chao.
\newblock Learning deep transformer models for machine translation.
\newblock In \emph{Proceedings of the 57th Annual Meeting of the Association for Computational Linguistics}, pp.\  1810--1822, 2019.

\bibitem[Xiao et~al.(2025)Xiao, Sang, Zhi, Liu, Yan, Luo, and Yuan]{xiao2025coap}
Jinqi Xiao, Shen Sang, Tiancheng Zhi, Jing Liu, Qing Yan, Linjie Luo, and Bo~Yuan.
\newblock Coap: Memory-efficient training with correlation-aware gradient projection.
\newblock In \emph{Proceedings of the IEEE/CVF Conference on Computer Vision and Pattern Recognition}, pp.\  30116--30126, 2025.

\bibitem[Yang et~al.(2024)Yang, Leng, Guo, Zhao, Nakada, Zhang, Yao, and Chen]{yang2024s}
Xinyu Yang, Jixuan Leng, Geyang Guo, Jiawei Zhao, Ryumei Nakada, Linjun Zhang, Huaxiu Yao, and Beidi Chen.
\newblock S$^2$ft: Efficient, scalable and generalizable llm fine-tuning by structured sparsity.
\newblock \emph{Advances in Neural Information Processing Systems}, 37:\penalty0 59912--59947, 2024.

\bibitem[Yu \& Wu(2023)Yu and Wu]{yu2023compressing}
Hao Yu and Jianxin Wu.
\newblock Compressing transformers: Features are low-rank, but weights are not!
\newblock In \emph{Proceedings of the 37th AAAI Conference on Artificial Intelligence}, pp.\  11007--11015, 2023.

\bibitem[Yuan et~al.(2023)Yuan, Shang, Song, Wu, Yan, and Sun]{yuan2023asvd}
Zhihang Yuan, Yuzhang Shang, Yue Song, Qiang Wu, Yan Yan, and Guangyu Sun.
\newblock Asvd: Activation-aware singular value decomposition for compressing large language models.
\newblock \emph{arXiv preprint arXiv:2312.05821}, 2023.

\bibitem[Zhang \& Sennrich(2019)Zhang and Sennrich]{zhang2019root}
Biao Zhang and Rico Sennrich.
\newblock Root mean square layer normalization.
\newblock In \emph{Advances in Neural Information Processing Systems}, volume~32, pp.\  12381--12392, 2019.

\bibitem[Zhang et~al.(2025)Zhang, Yin, Wang, Liu, Zhang, Shrivastava, Yang, and Braverman]{zhang2025i3s}
Haochen Zhang, Junze Yin, Guanchu Wang, Zirui Liu, Tianyi Zhang, Anshumali Shrivastava, Lin Yang, and Vladimir Braverman.
\newblock I3s: Importance sampling subspace selection for low-rank optimization in llm pretraining.
\newblock \emph{arXiv preprint arXiv:2502.05790}, 2025.

\bibitem[Zhang et~al.(2023)Zhang, Zhang, Shi, Chu, and Li]{zhang2023lora}
Longteng Zhang, Lin Zhang, Shaohuai Shi, Xiaowen Chu, and Bo~Li.
\newblock Lora-fa: Memory-efficient low-rank adaptation for large language models fine-tuning.
\newblock \emph{arXiv preprint arXiv:2308.03303}, 2023.

\bibitem[Zhang et~al.(2024)Zhang, Lou, Ying, Yang, and Zhou]{zhang2024efficient}
Luoming Zhang, Zhenyu Lou, Yangwei Ying, Cheng Yang, and Hong Zhou.
\newblock Efficient fine-tuning of large language models via a low-rank gradient estimator.
\newblock \emph{Applied Sciences}, 15:\penalty0 82, 2024.

\bibitem[Zhao et~al.(2025)Zhao, Xie, Fang, and Lin]{zhao2025separate}
Hanzhen Zhao, Xingyu Xie, Cong Fang, and Zhouchen Lin.
\newblock Separate: A simple low-rank projection for gradient compression in modern large-scale model training process.
\newblock In \emph{The 13rd International Conference on Learning Representations}, 2025.

\bibitem[Zhao et~al.(2024)Zhao, Zhang, Chen, Wang, Anandkumar, and Tian]{zhao2024galore}
Jiawei Zhao, Zhenyu Zhang, Beidi Chen, Zhangyang Wang, Anima Anandkumar, and Yuandong Tian.
\newblock Galore: Memory-efficient llm training by gradient low-rank projection.
\newblock In \emph{Proceedings of the 41st International Conference on Machine Learning}, pp.\  61121--61143, 2024.

\bibitem[Zhu et~al.(2024)Zhu, Zhang, Cong, Liu, Park, Chandra, Long, Pan, Wang, and Lee]{zhu2024apollo}
Hanqing Zhu, Zhenyu Zhang, Wenyan Cong, Xi~Liu, Sem Park, Vikas Chandra, Bo~Long, David~Z Pan, Zhangyang Wang, and Jinwon Lee.
\newblock Apollo: Sgd-like memory, adamw-level performance.
\newblock \emph{arXiv preprint arXiv:2412.05270}, 2024.

\end{thebibliography}
